\pdfoutput=1
\documentclass[11pt]{article}

\usepackage{EMNLP2023}

\usepackage{times}
\usepackage{latexsym}

\usepackage[T1]{fontenc}
\usepackage[utf8]{inputenc}

\usepackage{microtype}

\usepackage{inconsolata}

\usepackage{multirow}
\usepackage{graphicx}
\usepackage{booktabs}
\usepackage{color}
\usepackage{colortbl}
\usepackage{xcolor}
\usepackage{bm}
\usepackage{soul}
\usepackage{amsfonts}
\usepackage{amsmath}
\usepackage{subcaption}
\usepackage{textcomp}
\newcommand{\astfootnote}[1]{
    \let\oldthefootnote=\thefootnote
    \setcounter{footnote}{1}
    \renewcommand{\thefootnote}{\fnsymbol{footnote}}
    \footnotetext{#1}
    \let\thefootnote=\oldthefootnote
}

\title{
Universal Domain Adaptation for Robust Handling of \\ Distributional Shifts in NLP
}

\author{
    Hyuhng Joon Kim$^1$, Hyunsoo Cho$^{1,2}$, Sang-Woo Lee$^{2,3,4}$, Junyeob Kim$^1$,\\
    \textbf{Choonghyun Park$^1$, Sang-goo Lee$^{1,5}$, Kang Min Yoo$^{1,2,3*}$, Taeuk Kim$^{6*}$} \\
    $^1$Seoul National University, $^2$NAVER Cloud, $^3$NAVER AI Lab,  \\
    $^4$KAIST, $^5$IntelliSys, $^6$Hanyang University \\
    \texttt{\{heyjoonkim,johyunsoo,juny116,pch330,sglee\}@europa.snu.ac.kr} \\
    \texttt{\{sang.woo.lee,kangmin.yoo\}@navercorp.com} \\
    \texttt{kimtaeuk@hanyang.ac.kr}
}

\begin{document}
\maketitle
\begin{abstract}

When deploying machine learning systems to the wild, it is highly desirable for them to effectively leverage prior knowledge to the unfamiliar domain while also firing alarms to anomalous inputs.
In order to address these requirements, Universal Domain Adaptation (UniDA) has emerged as a novel research area in computer vision, focusing on achieving both adaptation ability and robustness (i.e., the ability to detect out-of-distribution samples).
While UniDA has led significant progress in computer vision, its application on language input still needs to be explored despite its feasibility.
In this paper, we propose a comprehensive benchmark for natural language that offers thorough viewpoints of the model's generalizability and robustness.
Our benchmark encompasses multiple datasets with varying difficulty levels and characteristics, including temporal shifts and diverse domains.
On top of our testbed, we validate existing UniDA methods from computer vision and state-of-the-art domain adaptation techniques from NLP literature, yielding valuable findings:
We observe that UniDA methods originally designed for image input can be effectively transferred to the natural language domain while also underscoring the effect of adaptation difficulty in determining the model's performance.

\end{abstract}
\astfootnote{Co-corresponding authors.}
\section{Introduction}
\label{sec:introduction}

\begin{figure}[t]
    \begin{center}
        \includegraphics[width=0.99\columnwidth]{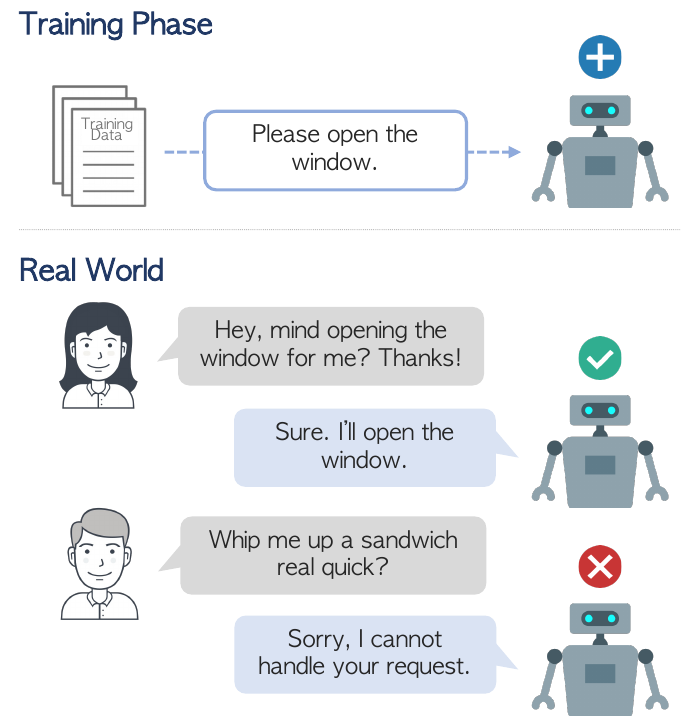}
          \caption{
          % Two research questions, i.e., domain adaptation and OOD detection, can be raised when a model faces distributional shifts. 
          The model trained with formal language (source domain) will likely face spoken language (target domain) in the real world.
          The model is expected to properly handle such transferable input despite the distributional shift. (middle)
          At the same time, the model should discern unprocessable inputs (bottom) from the target domain. 
          % Universal domain adaptation aims to properly handle transferable inputs (middle),
          % and discern unprocessable inputs (bottom)
          % from the target domain.
          }
          \label{front_image}
    \end{center}
\end{figure}

% revision

Deep learning models demonstrate satisfactory performance when tested on data from the training distribution. 
However, real-world inputs encounter novel data ceaselessly that deviate from the trained distribution, commonly known as distributional shift.
When confronted with such inputs, machine learning models frequently struggle to differentiate them from regular input.
Consequently, they face challenges in adapting their previously acquired knowledge to the new data distribution, resulting in a significant drop in performance \citep{ribeiro-etal-2020-beyond,pmlr-v119-miller20a,hendrycks-etal-2020-pretrained}.
The aforementioned phenomenon represents a longstanding challenge within the machine learning community, wherein even recent cutting-edge language models \citep{openai2023gpt4, touvron2023llama, chowdhery2022palm, NEURIPS2020_1457c0d6}
 do not serve as an exception to this predicament \citep{wang2013robustness}.

In response to these challenges, existing literature proposes two distinct approaches.
The first approach, known as Domain Adaptation (DA) \citep{blitzer-etal-2006-domain,ganin2016domain,Karouzos2021UDALMUD,wu-shi-2022-adversarial}, 
endeavors to establish alignment between a new set of data from an unknown distribution and the model's prior knowledge distribution.
The objective is to enhance the model's generalization capability and reduce performance drop springing from the distributional shift.
In parallel, a distinct line of work, referred to as out-of-distribution (OOD) detection \citep{Aggarwal2017, hendrycks2017a, hendrycks2018deep, ijcai2021p198}, focuses on discerning inputs originating from dissimilar distributions.
They opt to circumvent potential risks or disruptions arising from shifted inputs, thereby enriching system robustness and resilience.

While both approaches offer unique advantages addressing specific distributional shifts, integrating their merits could substantially enhance robustness.
In pursuit of this objective, a novel field called \textbf{Universal Domain Adaptation (UniDA)} \citep{UDA_2019_CVPR} has emerged, aiming to harness the synergies of both OOD detection and DA when confronted with distributional shifts.
UniDA leverages the best of the two worlds and offers comprehensive perspectives that integrate the merits of these two research areas.
The essence of UniDA lies in measuring the uncertainty of the data from the shifted distribution precisely. 
Then, we can enhance the model’s transferability by distinguishing portions of low-uncertainty inputs that can be adequately handled with the current model’s knowledge. 
Simultaneously, we enrich the robustness of the model to OOD inputs by discerning the remaining samples that cannot be processed normally.
However, distinguishing between these inputs and properly processing them becomes increasingly challenging without explicit supervision.

Despite the versatility of UniDA, this topic has yet to be explored in the Natural Language Processing (NLP) literature.
As a cornerstone in enhancing reliability against distributional shifts in NLP, we introduce a testbed for evaluating the model's robustness in a holistic view.
First, we construct various adaptation scenarios in NLP, utilizing an array of thoughtfully selected datasets.
To discern the degree to which our proposed datasets incorporate the various degree of challenges in UniDA, we define two novel metrics:
\textbf{Performance Drop Rate (PDR)} and \textbf{Distinction Difficulty Score (DDS)}. 
Using these metrics, We verify that our testbed captures a broad spectrum of distributional shifts.
Finally, based on the suggested setting, we systematically compare several UniDA methods inherently designed for the task, 
against heuristic combinations of previous approaches for the parts of the problem, i.e., OOD detection and DA.

Our empirical results show that UniDA methods are fully transferable in the NLP domain 
and can robustly respond to various degrees of shift.
Moreover, we find out that the adaptation difficulty notably affects the performance of the methods.
In certain circumstances, DA methods display comparable or even better performance. 
We release our dataset, encouraging future research on UniDA in NLP to foster the development of more resilient and domain-specific strategies.\footnote{The dataset is available at \url{https://github.com/heyjoonkim/universal_domain_adaptation_for_nlp}}

% \subsection{UniDA}
\section{Universal Domain Adaptation}
\subsection{Problem Formulation}
Distributional shift refers to a situation where the joint distribution $P$ estimated from the training dataset fails to adequately represent the wide range of diverse and complex test inputs.
More formally, a distributional shift arises when the test input $\bm{x}_{\text{test}}$ originates from a distant distribution $Q$, which is not effectively encompassed by the current trained distribution $P$.

One of the most prevalent research areas addressing this distributional shift includes OOD detection and DA.
OOD detection aims to strictly detect all inputs from $Q$ to enhance the model's reliability.
Although distribution $Q$ demonstrates a discernibly different distribution from the distribution $P$, 
the trained model can still transfer a subset of instances from $Q$, overcoming the inherent discrepancy between $P$ and $Q$. 
This particular capability serves as a fundamental motivation underlying the pursuit of DA.
UniDA endeavours to integrate the merits of both fields, thereby enhancing both the generalizability and reliability of the model.
Specifically, let us divide the target distribution $Q$ into disjoint subsets $H$, which share the same label space with source distribution $P$ and its complement $I$ ($Q=H \cup I$).
The objective of UniDA is to enrich the robustness of the model by flexibly transferring existing knowledge to \textit{transferable} samples from $H$ while firing alarms to \textit{unknown} samples from $I$.

%%%%%%%%%%%%%%%%%%%%%%%%%%%%%%%%%%%%%%%%%%%%%%%%%%%%%%%%%%
%%%%%%%%%%%%%%%%%%%%%%%%%%%%%%%%%%%%%%%%%%%%%%%%%%%%%%%%%%

\subsection{Challenges in UniDA}
UniDA models should be capable of accurately capturing the underlying reasons behind the shift, thereby enabling the discrimination between \textit{transferable} samples and \textit{unknown} samples.
Among the diverse categories of causes, the \textbf{domain gap} and the \textbf{category gap} \citep{UDA_2019_CVPR} emerge as pivotal factors, each exerting a substantial impact on the overall complexity of the UniDA problem. 
While these concepts have previously been defined in a rather vague manner, 
we deduced the necessity for a more explicit definition.
Thus, we set forth to redefine the concepts of domain and category gap more explicitly.

Domain gap refers to the performance drop 
when a model trained on $P$ fails to correctly process \textit{transferable} inputs
due to the fundamental discrepancy between $P$ and $H$, i.e., a domain shift.
A dataset with a higher domain gap amplifies the problem's difficulty as the trained model becomes more susceptible to misaligning \textit{transferable} samples.

A category shift, characterized by the disparity in the class sets considered by $P$ and $I$, causes a category gap.
Category gap represents the performance drop that arises for inputs from $I$, 
which cannot be processed properly due to differing class sets between $P$ and $I$, 
which are erroneously handled.
A larger category gap makes distinguishing \textit{unknown} samples from \textit{transferable} samples harder, thereby worsening the robustness of the model.

From the perspective of addressing the domain gap and category gap, 
the main goal of UniDA is to minimize both gaps simultaneously. 
This aims to ensure \textit{transferable} samples properly align with the source domain for adequate processing,
while handling \textit{unknown} samples as unprocessable exceptions.

%%%%%%%%%%%%%%%%%%%%%%%%%%%%%%%%%%%%%%%%%%
%%%%%%%%%%%%%%%%%%%%%%%%%%%%%%%%%%%%%%%%%%
\section{Testbed Design}
% \subsection{Motivation}

The primary objective of our research is to construct a comprehensive benchmark dataset that effectively captures the viewpoint of UniDA.
To accomplish our objective, we attempt to create a diverse dataset that encompasses a range of difficulty levels and characteristics, such as domains, sentiment, or temporal change.
These variations are the fundamental elements that can significantly influence the overall performance.

Specifically, we initially select datasets from multiple practical domains
and approximate the adaptation difficulty by quantifying different shifts with our newly proposed metrics.
In the following subsections, we provide an in-depth explanation of our dataset along with the analysis of our benchmarks.

%%%%%%%%%%%%%%%%%%%%%%%%%%%%%%%%%%%%%%%%%%
%%%%%%%%%%%%%%%%%%%%%%%%%%%%%%%%%%%%%%%%%%

% lacks a clear and quantitative approach for measuring domain and category gaps. 

\subsection{Quantifying Different Shifts}
As the extent of both domain and category gaps significantly influences the overall adaptation complexity, 
it is essential to quantify these gaps when designing the dataset for evaluation.
Unfortunately, existing literature has not devised a clear-cut and quantitative measure for assessing domain and category gaps.
Therefore, we endeavoured to define measures that can aptly approximate the two types of gaps.

\textbf{Performance Drop Rate (PDR)} 
measures the degree of domain gap by assessing the absolute difficulty of the dataset itself and the performance drop caused by the shift from $P$ to $H$. 
Specifically, we fine-tune \textit{bert-base-uncased} on the source train set and evaluate its test set accuracy $acc_s$ from the same distribution. 
Leveraging the same model trained on the source domain, we then measure the accuracy of the target test set $acc_t$. 
We measure the performance degradation caused by the distributional shift by measuring $acc_s-acc_t$.
Since the significance of the performance drop may vary depending on the source performance, 
we normalize the result with the source performance and measure the proportion of the performance degradation. 
A more significant drop rate indicates a greater decline in performance, considering the source domain performance.
Formally, PDR for a source domain $s$ and a target domain $t$ can be measured as follows:
\begin{equation}
  \text{PDR}_{s,t} = 100 \times \frac{acc_s - acc_t}{acc_s}
\end{equation}

\textbf{Distinction Difficulty Score (DDS)}
is measured to estimate the difficulty of distinguishing between $H$ and $I$, which, in other words, measures the difficulty of handling the category shift.
We utilized the same model trained on the source domain and extracted the \texttt{[CLS]} representations of the source inputs.
We estimated the source distribution, 
assuming the extracted representations follow the multivariate normal distribution.
We then extracted \texttt{[CLS]} representations of target distribution inputs from the same model
and measured the Mahalanobis distance between the source distribution.
Using the distance, we measured the \textit{Area Under the ROC Curve} (AUC) as a metric for discerning $I$ and $H$.
AUC values closer to 1 indicate the ease of discerning \textit{unknown} inputs from the \textit{transferable} inputs.
Since we focus on quantifying the difficulty in distinguishing the two, 
we subtract the AUC from 1 to derive our final measure of interest.
For the source domain $s$, the target domain $t$, 
and AUC as $\text{AUC}_{s,t}$, DDS can be measured as:
\begin{equation}
  \text{DDS}_{s,t} = 100 \times (1 - \text{AUC}_{s,t})
\end{equation}

%%%%%%%%%%%%%%%%%%%%%%%%%%%%%%%%%%%%%%%%%%
%%%%%%%%%%%%%%%%%%%%%%%%%%%%%%%%%%%%%%%%%%

\subsection{Implementation of Different Shifts}
To construct a representative testbed for UniDA, it is essential to illustrate domain and category shifts. 
To exhibit domain shift, we delineated domains from various perspectives. 
This involves explicit factors such as temporal or sentiment and implicit definitions based on the composition of the class set. 
Detailed formation of domains for each dataset is stipulated in Section \ref{sec:dataset_details}.

To establish category shifts, the source and the target domain must have a set of common classes $C$ and a set of their own private classes, $\Bar{C}_{s}$ and $\Bar{C}_{t}$, respectively.
We followed previous works \citep{UDA_2019_CVPR,Fu2020LearningTD} by sorting the class name in alphabetical order and selecting the first $|C|$ classes as common, the subsequent $|\Bar{C}_{s}|$ as source private, and the rest as target private classes. The class splits for each dataset are stated as $|C|/|\Bar{C}_{s}|/|\Bar{C}_{t}|$ in the main experiments.

%%%%%%%%%%%%%%%%%%%%%%%%%%%%%%%%%%%%%%%%%%
%%%%%%%%%%%%%%%%%%%%%%%%%%%%%%%%%%%%%%%%%%

\subsection{Dataset Details}
\label{sec:dataset_details}
We focused on text classification tasks for our experiments.
Four datasets were selected from multiple widely used classification domains in NLP, such as
topic classification, sentiment analysis, and intent classification.
We reformulated the datasets so that our testbed could cover diverse adaptation scenarios.

\textbf{Huffpost News Topic Classification (Huffpost)} \citep{misra2022news} 
contains Huffpost news headlines spanning from 2012 to 2022.
The task is to classify news categories given the headlines.
Using the temporal information additionally provided, 
we split the dataset year-wise from 2012 to 2017, treating each year as a distinct domain.
We selected the year 2012 as the source domain, with the subsequent years assigned as the target domains,
creating 5 different levels of temporal shifts.

\textbf{Multilingual Amazon Reviews Corpus (Amazon)} \citep{keung-etal-2020-multilingual}
includes product reviews that are commonly used to predict star ratings based on the review, 
and additional product information is provided for each review.
We have revised the task to predict the product information given the reviews
and utilized the star ratings to define sentiment domains.
Reviews with a star rating of 1 or 2 are grouped as negative sentiment, 
and those with a rating of 4 or 5 are categorized as positive.
We exclude 3-star reviews, considering them neutral.

\textbf{MASSIVE} \citep{fitzgerald2022massive}
is a hierarchical dataset for intent classification. 
The dataset consists of 18 first-level and 60 second-level classes.
Each domain is defined as a set of classes, including private classes exclusive to a specific domain
and common classes shared across domains.
We divided the common first-level class into two parts based on second-level classes to simulate domain discrepancy. 
The half of the divided common class represents the source domain while the other half represents the target domain.
We assume that the second-level classes within the same first-level class share a common feature and thus can be adapted.

\textbf{CLINC-150} \citep{larson-etal-2019-evaluation}
is widely used for intent classification in OOD detection.
The dataset consists of 150 second-level classes
over 10 first level-classes and a single out-of-scope class.
The domain is defined in the same way as MASSIVE.

%%%%%%%%%%%%%%%%%%%%%%%%%%%%%%%%%%%%%%%%%%
%%%%%%%%%%%%%%%%%%%%%%%%%%%%%%%%%%%%%%%%%%

\begin{table}
    \centering 
    \resizebox{0.99 \linewidth}{!}{

\begin{tabular}{c|cccc}
\toprule
Dataset & \begin{tabular}[c]{@{}c@{}}Huffpost\\ (2013)\end{tabular} & \begin{tabular}[c]{@{}c@{}}Huffpost\\ (2014)\end{tabular} & \begin{tabular}[c]{@{}c@{}}Huffpost\\ (2015)\end{tabular} & \begin{tabular}[c]{@{}c@{}}Huffpost\\ (2016)\end{tabular} \\ \midrule
% Source Acc. & 88.82 & 88.82 & 88.82 & 88.82 \\
% Target Acc. & 81.83 & 70.44 & 70.64 & 66.61 \\
% Difference & 6.99 & 18.38 & 18.18 & 22.21 \\
PDR & 7.87 & 20.69 & 20.47 & 25.00 \\ 
DDS & 19.67 & 27.79 & 25.71 & 31.19 \\ \midrule
Dataset & \begin{tabular}[c]{@{}c@{}}Huffpost\\ (2017)\end{tabular} & CLINC-150 & MASSIVE & Amazon \\ \midrule
% Source Acc. & 88.82 & 98.75 & 98.01 & 61.40 \\
% Target Acc. & 58.96 & 87.30 & 62.58 & 51.91 \\
% Difference & 29.86 & 11.45 & 35.42 & 9.49 \\
PDR & 33.62 & 11.59 & \textbf{36.14} & 15.45 \\ 
DDS & 31.74 & 13.45 & 23.92 & \textbf{36.04} \\ \bottomrule
\end{tabular}

% \begin{tabular}{c|cccc}
% \toprule
% Dataset & \begin{tabular}[c]{@{}c@{}}Huffpost\\ (2013)\end{tabular} & \begin{tabular}[c]{@{}c@{}}Huffpost\\ (2014)\end{tabular} & \begin{tabular}[c]{@{}c@{}}Huffpost\\ (2015)\end{tabular} & \begin{tabular}[c]{@{}c@{}}Huffpost\\ (2016)\end{tabular} \\ \midrule
% Source Acc. & 88.82 & 88.82 & 88.82 & 88.82 \\
% Target Acc. & 81.83 & 70.44 & 70.64 & 66.61 \\
% Difference & 6.99 & 18.38 & 18.18 & 22.21 \\
% PDR & 7.87 & 20.69 & 20.47 & 25.00 \\ \midrule
% DDS & 19.67 & 27.79 & 25.71 & 31.19 \\ \midrule
% Dataset & \begin{tabular}[c]{@{}c@{}}Huffpost\\ (2017)\end{tabular} & CLINC-150 & MASSIVE & Amazon \\ \midrule
% Source Acc. & 88.82 & 98.75 & 98.01 & 61.40 \\
% Target Acc. & 58.96 & 87.30 & 62.58 & 51.91 \\
% Difference & 29.86 & 11.45 & 35.42 & 9.49 \\
% PDR & 33.62 & 11.59 & \textbf{36.14} & 15.45 \\ \midrule
% DDS & 31.74 & 13.45 & 23.92 & \textbf{36.04} \\ \bottomrule
% \end{tabular}

    }
    \caption{
        \label{table:shift_measures}
        % The results of the approximate measure of domain and category gap.
        PDR and DDS values of each datasets.
        The \textbf{largest value} of PDR and DDS is highlighted in bold.
    }
\end{table}

%%%%%%%%%%%%%%%%%%%%%%%%%%%%%%%%%%%%%%%%%%
%%%%%%%%%%%%%%%%%%%%%%%%%%%%%%%%%%%%%%%%%%
\subsection{Dataset Analysis}
\label{sec:dataset_analysis}
In this section, we intend to validate whether our testbed successfully demonstrates diverse adaptation difficulties, aligning with our original motivation. 
We assess adaptation difficulty from domain and category gap perspectives, each approximated by PDR and DDS, respectively.

The results of PDR and DDS are reported in Table \ref{table:shift_measures}.
The result shows diverse PDR values ranging from 7 to 36 points, 
indicating various degrees of domain gap across the datasets.
MASSIVE measured the most considerable domain gap, 
while Huffpost (2013) demonstrated the most negligible domain gap among the datasets.
Additionally, our testbed covers a wide range of category gaps, indicated by the broad spectrum of DDS values.
Specifically, Amazon exhibits a significantly high DDS value, representing an extremely challenging scenario
of differentiating the \textit{unknown} samples from the \textit{transferable} samples.

\begin{figure}[t]
    \begin{center}
        \includegraphics[width=0.99\columnwidth]{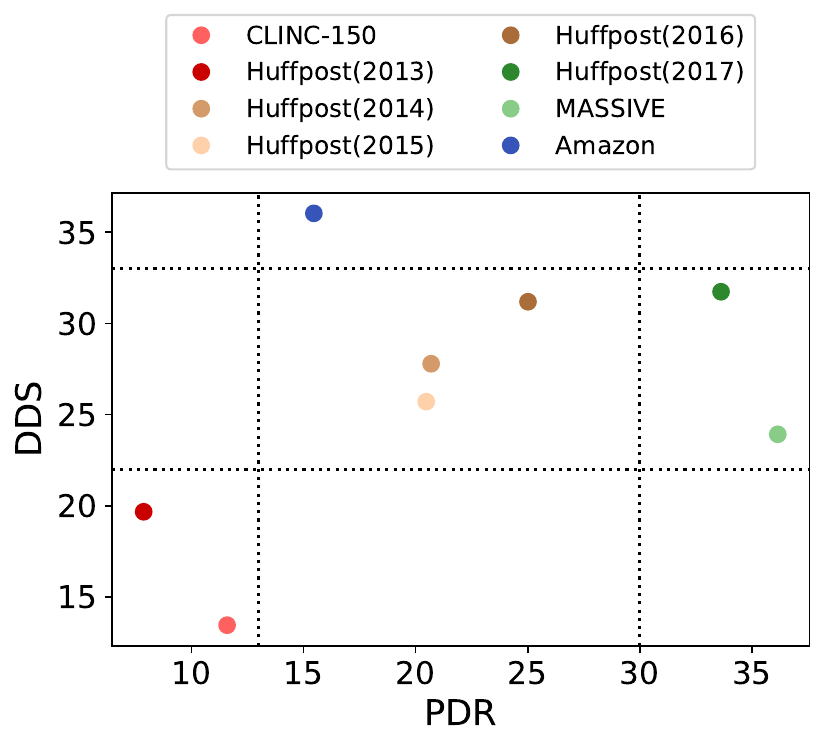}
          \caption{
          The visualization of adaptation difficulty for each dataset 
          in terms of domain gap (PDR) and category gap (DDS). 
          We categorized the dataset into 4 distinct groups based on the adaptation complexity.
          Best viewed in color. 
          }
          \label{fig:shift_plot}
    \end{center}
\end{figure}

We consolidate the two indicators to measure the coverage of different adaptation complexity of our proposed testbed. 
We visualized the datasets considering the PDR and DDS; metrics on the respective axes.
Figure \ref{fig:shift_plot} is the visualization of the adaptation complexity for each dataset factored by PDR and DDS.
We grouped the datasets into four distinct clusters based on the plotted distribution.
Datasets closer to the lower-left corner, CLINC-150 and Huffpost (2013), are easy adaptation scenarios with minor domain and category gaps.
Datasets plotted in the center, Huffpost (2014, 2015, 2016), presents moderate difficulty.
Amazon suffers from a category gap significantly, 
while Huffpost (2017) and MASSIVE demonstrate a notable domain gap, yielding high adaptation complexity.
The results validate that our testbed embodies a diverse range of adaptation difficulties as intended.

% \section{Dataset Design}
% \label{sec:dataset_analysis}
% \input{Sections/03. dataset_analysis}

\label{sec:setting}
\section{Experimental Setting}

\subsection{Compared Methods}
We compare several domain adaptation methods on our proposed testbed.
We selected two previous state-of-the-art closed-set Domain Adaptation (CDA) methods, UDALM \citep{Karouzos2021UDALMUD} and AdSPT \citep{wu-shi-2022-adversarial},
under the assumption that all the inputs from the target domain are \textit{transferable} without considering \textit{unknown} classes.
Two previous state-of-the-art UniDA methods were selected, 
OVANet \citep{saito2021ovanet} and UniOT \citep{chang2022unified},
which are fully optimized to handle UniDA scenarios in the vision domain.
We also conducted experiments with additional baseline methods such as DANN \citep{Ganin2016DomainAdversarialTO}, UAN \citep{UDA_2019_CVPR}, and CMU \citep{Fu2020LearningTD}.
However, the performance was subpar compared to the selected methods, exhibiting a similar tendency. 
Hence, we report the additional results in Appendix \ref{sec:appendix_full_experimental_results}.
For the backbone of all the methods, we utilized \textit{bert-base-uncased} \citep{devlin-etal-2019-bert} and used the \texttt{[CLS]} representation as the input feature.
Implementation details are stipulated in Appendix \ref{sec:implementation_details}.

%%%%%%%%%%%%%%%%%%%%%%%%%%%%%%%%%%%%%%%%%%%%%%%%%%%%%%%%%%%%%%%%%%%%%%%%%%%%%%%%%%%%%%%%%%%%%%%%%%%%%%%%

\subsection{Thresholding Method}
Since CDA methods are not designed to handle \textit{unknown} inputs, 
additional techniques are required to discern them.
A straightforward yet intuitive approach to detecting \textit{unknown} inputs is applying a threshold for the output of the scoring function.
The scoring function reflects the appropriateness of the input based on the extracted representation.
If the output of the scoring function falls below the threshold,
the instance is classified as \textit{unknown}.
We sequentially apply thresholding after the adaptation process.\footnote{
In the case of applying the thresholding first, all the inputs from the target domain would be classified as OOD. 
If all the target inputs were classified as OOD, the criteria for discerning \textit{transferable} and \textit{unknown} inputs become inherently unclear. 
Therefore, we have only considered scenarios where thresholding is applied after the adaptation.
}
Formally, for an input $x$, categorical prediction $ \hat{y} $, threshold value $w$, and a scoring function $ f_{score}$, the final prediction is made as:
\begin{equation}
  y(x)=\begin{cases}
    \text{argmax}(\hat{y}), & \text{if $f_{score}(x) > w$}\\
    \texttt{unknown}, & \text{otherwise}.
  \end{cases}
\end{equation}

We utilize Maximum Softmax Probability (MSP) as the scoring function\footnote{In addition to MSP, we have applied various scoring functions such as cosine similarity and Mahalanobis distance, but using MSP achieved the best performance.
Results for other thresholding methods have been included in the Appendix \ref{sec:appendix_full_experimental_results}.} \citep{hendrycks2017a}.
Following the OOD detection literature, the value at the point of 95\% from the sorted score values was selected as the threshold.

%%%%%%%%%%%%%%%%%%%%%%%%%%%%%%%%%%%%%%%%%%%%%%%%%%%%%%%%%%%%%%%%%%%%%%%%%%%%%%%%%%%%%%%%%%%%%%%%%%%%%%%%

\subsection{Evaluation Protocol}
The goal of UniDA is to properly process the \textit{transferable} inputs and detect the \textit{unknown} inputs simultaneously, 
consequently making both the \textit{transferable} and the \textit{unknown} accuracies crucial metrics.
We applied H-score \citep{Fu2020LearningTD} as the primary evaluation metric to integrate both evaluation metrics.
H-score is the harmonic mean between the accuracy of common class $acc_{C}$ and \textit{unknown} class $ acc_{\bar{C}_t} $,
where $acc_{C}$ is the accuracy over the common class set $C$ and $acc_{\bar{C}_t}$ is the accuracy predicting the \textit{unknown} class.
The model with a high H-score is considered robust in the UniDA setting,
indicating its proficiency in both adaptation (high $acc_{C}$) 
and OOD detection (high $acc_{\bar{C}_t}$).
Formally, the H-score can be defined as :
\begin{equation}
    H_{score} = 2 \cdot \frac{acc_{C} \cdot acc_{\bar{C}_t}}{acc_{C} + acc_{\bar{C}_t}}
\end{equation}

Although the H-score serves as an effective evaluation criterion for UniDA, 
we also report $acc_{C}$ and $acc_{\bar{C}_t}$ to provide a comprehensive assessment. 
We report the averaged results and standard deviations over four runs for all experiments.

\label{sec:results}
\section{Experimental Results}

\begin{figure}[t]
    \begin{center}
        \includegraphics[width=0.99\columnwidth]{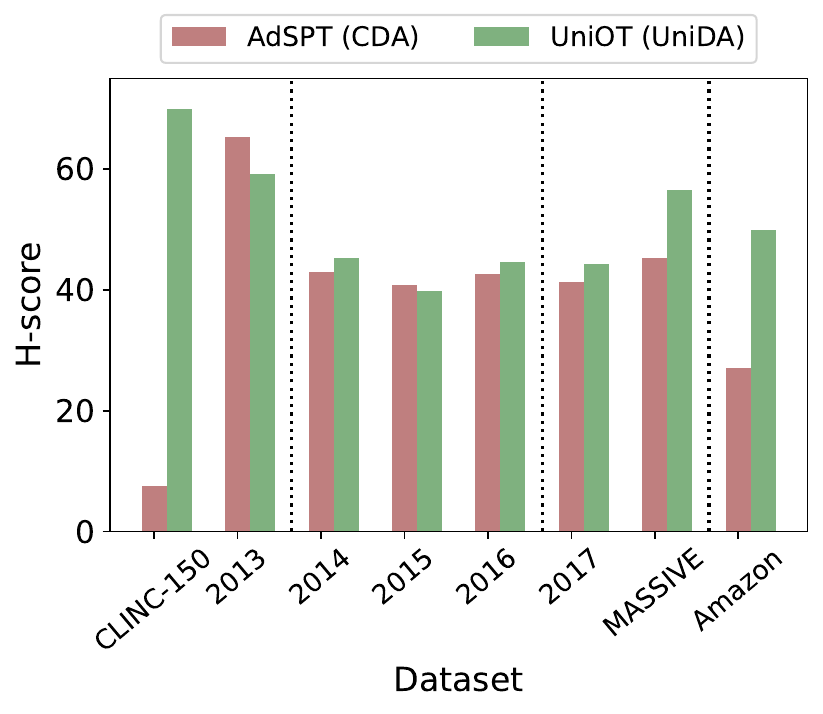}
          \caption{
          H-score results of AdSPT and UniOT on all the datasets.
          The preferred method varies depending on the adaptation complexity.
          }
          \label{fig:overal_results}
    \end{center}
\end{figure}

\begin{table}[t]
    \centering
    \resizebox{0.99 \linewidth}{!}{

\begin{tabular}{cccc}
\toprule
\multicolumn{4}{c}{Huffpost (2012 $\rightarrow$ 2013) (3 / 4 / 4)} \\ \midrule
\multicolumn{1}{c|}{Method} & $acc_{C}$ & $ acc_{\bar{C}_t} $ & H-score \\ \midrule
\multicolumn{1}{c|}{UDALM} & 52.74 {\footnotesize $\pm 2.92$} & 58.55 {\footnotesize $\pm 6.03$} & 55.15 {\footnotesize $\pm 2.37$} \\
\multicolumn{1}{c|}{AdSPT} & 55.05 {\footnotesize $\pm 2.11$} & 80.66 {\footnotesize $\pm 2.99$} & \textbf{65.38} {\footnotesize $\pm 1.00$} \\
\midrule
\multicolumn{1}{c|}{OVANet} & 65.11  {\footnotesize $\pm 0.60$} & 24.91 {\footnotesize $\pm 6.75$} & 35.64 {\footnotesize $\pm 6.70$} \\
\multicolumn{1}{c|}{UniOT} & 53.76 {\footnotesize $\pm 1.10$} & 65.86 {\footnotesize $\pm 3.57$} & \underline{59.14} {\footnotesize $\pm 1.14$} \\ \midrule
\multicolumn{4}{c}{CLINC-150 (4 / 3 / 3)} \\ \midrule
\multicolumn{1}{c|}{Method} & $acc_{C}$ & $ acc_{\bar{C}_t} $ & H-score \\ \midrule
\multicolumn{1}{c|}{UDALM} & 74.92 {\footnotesize $\pm 6.11$} & 69.91 {\footnotesize $\pm 13.98$} & \textbf{71.28} {\footnotesize $\pm 4.62$} \\
\multicolumn{1}{c|}{AdSPT} & 3.96 {\footnotesize $\pm 2.17$} & 98.69 {\footnotesize $\pm 0.36$} & 7.55 {\footnotesize $\pm 3.95$} \\
\midrule
\multicolumn{1}{c|}{OVANet} & 83.49  {\footnotesize $\pm 0.62$} & 31.24 {\footnotesize $\pm 1.70$} & 45.45 {\footnotesize $\pm 1.88$} \\
\multicolumn{1}{c|}{UniOT} & 64.14 {\footnotesize $\pm 9.14$} & 77.36 {\footnotesize $\pm 3.94$} & \underline{69.88} {\footnotesize $\pm 6.11$} \\ \bottomrule
\end{tabular}

    }
    \caption{
        \label{table:easy_results}
        Experimental results on CLINC-150 and Huffpost (2013), which is a relatively easy adaptation scenario.
        The \textbf{best method} with the highest H-score is in bold, and the \underline{second-best method} is underlined.
    }
\end{table}

\begin{table}[t]
    \centering
    \resizebox{0.99 \linewidth}{!}{

\begin{tabular}{cccc}
\toprule
\multicolumn{4}{c}{Huffpost (2012 $\rightarrow$ 2014) (3 / 4 / 4)} \\ \midrule
\multicolumn{1}{c|}{Method} & $acc_{C}$ & $ acc_{\bar{C}_t} $ & H-score \\ \midrule
\multicolumn{1}{c|}{UDALM} & 29.90 {\footnotesize $\pm 1.89$} & 50.00 {\footnotesize $\pm 17.49$} & 34.14 {\footnotesize $\pm 5.07$} \\
\multicolumn{1}{c|}{AdSPT} & 29.64 {\footnotesize $\pm 4.51$} & 80.03 {\footnotesize $\pm 6.09$} & \underline{42.93} {\footnotesize $\pm 3.41$} \\
\midrule
\multicolumn{1}{c|}{OVANet} & 45.85 {\footnotesize $\pm 2.62$} & 33.96 {\footnotesize $\pm 6.78$} & 36.02 {\footnotesize $\pm 3.95$} \\
\multicolumn{1}{c|}{UniOT} & 33.49 {\footnotesize $\pm 4.79$} & 71.44 {\footnotesize $\pm 6.19$} & \textbf{45.23} {\footnotesize $\pm 3.60$} \\ \midrule
\multicolumn{4}{c}{Huffpost (2012 $\rightarrow$ 2015) (3 / 4 / 4)} \\ \midrule
\multicolumn{1}{c|}{Method} & $acc_{C}$ & $ acc_{\bar{C}_t} $ & H-score \\ \midrule
\multicolumn{1}{c|}{UDALM} & 25.15 {\footnotesize $\pm 2.60$} & 61.79 {\footnotesize $\pm 8.88$} & 35.56 {\footnotesize $\pm 3.15$} \\
\multicolumn{1}{c|}{AdSPT} & 29.08 {\footnotesize $\pm 5.39$} & 73.04 {\footnotesize $\pm 12.71$} & \textbf{40.78} {\footnotesize $\pm 3.42$} \\
\midrule
\multicolumn{1}{c|}{OVANet} & 43.91 {\footnotesize $\pm 2.09$} & 35.70 {\footnotesize $\pm 4.41$} & 39.21 {\footnotesize $\pm 2.43$} \\
\multicolumn{1}{c|}{UniOT} & 27.16 {\footnotesize $\pm 3.40$} & 75.50 {\footnotesize $\pm 2.68$} & \underline{39.82} {\footnotesize $\pm 3.39$} \\ \midrule
\multicolumn{4}{c}{Huffpost (2012 $\rightarrow$ 2016) (3 / 4 / 4)} \\ \midrule
\multicolumn{1}{c|}{Method} & $acc_{C}$ & $ acc_{\bar{C}_t} $ & H-score \\ \midrule
\multicolumn{1}{c|}{UDALM} & 29.14 {\footnotesize $\pm 1.59$} & 60.87 {\footnotesize $\pm 6.87$} & 39.28 {\footnotesize $\pm 1.44$} \\
\multicolumn{1}{c|}{AdSPT} & 28.84 {\footnotesize $\pm 3.57$} & 82.52 {\footnotesize $\pm 4.22$} & \underline{42.55} {\footnotesize $\pm 3.27$} \\
\midrule
\multicolumn{1}{c|}{OVANet} & 44.30 {\footnotesize $\pm 2.89$} & 33.49 {\footnotesize $\pm 4.79$} & 38.03 {\footnotesize $\pm 3.79$} \\
\multicolumn{1}{c|}{UniOT} & 33.09 {\footnotesize $\pm 3.62$} & 69.27 {\footnotesize $\pm 3.83$} & \textbf{44.60} {\footnotesize $\pm 2.85$} \\ \bottomrule
\end{tabular}

    }
    \caption{
        \label{table:mid_results}
        Experimental results on Huffpost (2014, 2015, 2016), which has a moderate complexity for adaptation.
        The \textbf{best method} with the highest H-score is in bold, and the \underline{second-best method} is underlined.
    }
\end{table}

\begin{table}[t]
    \centering
    \resizebox{0.99 \linewidth}{!}{

\begin{tabular}{cccc}
\toprule
\multicolumn{4}{c}{Amazon (11 / 10 / 10)} \\ \midrule
\multicolumn{1}{c|}{Method} & $acc_{C}$ & $ acc_{\bar{C}_t} $ & H-score \\ \midrule
\multicolumn{1}{c|}{UDALM} & 47.69 {\footnotesize $\pm 1.47$} & 15.10 {\footnotesize $\pm 2.30$} & 22.85 {\footnotesize $\pm 2.65$} \\
\multicolumn{1}{c|}{AdSPT} & 30.26 {\footnotesize $\pm 3.52$} & 30.93 {\footnotesize $\pm 26.14$} & 27.12 {\footnotesize $\pm 11.03$} \\
\midrule
\multicolumn{1}{c|}{OVANet} & 44.50 {\footnotesize $\pm 0.84$} & 42.49 {\footnotesize $\pm 3.95$} & \underline{43.38} {\footnotesize $\pm 1.83$} \\
\multicolumn{1}{c|}{UniOT} & 38.60 {\footnotesize $\pm 1.08$} & 70.98 {\footnotesize $\pm 7.76$} & \textbf{49.90} {\footnotesize $\pm 2.08$} \\ \bottomrule
\end{tabular}

    }
    \caption{
        \label{table:category_hard_results}
        Experimental results on Amazon, which has a significant influence of category gap.
        The \textbf{best method} with the highest H-score are in bold, and the \underline{second-best method} is underlined.
    }
\end{table}

\begin{table}[t]
    \centering
    \resizebox{0.99 \linewidth}{!}{

\begin{tabular}{cccc}
\toprule
\multicolumn{4}{c}{MASSIVE (8 / 5 / 5)} \\ \midrule
\multicolumn{1}{c|}{Method} & $acc_{C}$ & $ acc_{\bar{C}_t} $ & H-score \\ \midrule
\multicolumn{1}{c|}{UDALM} & 38.33 {\footnotesize $\pm 4.88$} & 41.29 {\footnotesize $\pm 15.14$} & 39.28 {\footnotesize $\pm 9.69$} \\
\multicolumn{1}{c|}{AdSPT} & 36.78 {\footnotesize $\pm 2.65$} & 59.87 {\footnotesize $\pm 13.47$} & 45.32 {\footnotesize $\pm 6.20$} \\
\midrule
\multicolumn{1}{c|}{OVANet} & 59.04 {\footnotesize $\pm 0.69$} & 39.28 {\footnotesize $\pm 2.14$} & \underline{47.12} {\footnotesize $\pm 1.47$} \\
\multicolumn{1}{c|}{UniOT} & 44.01 {\footnotesize $\pm 0.20$} & 79.19 {\footnotesize $\pm 3.52$} & \textbf{56.52} {\footnotesize $\pm 3.49$} \\ \midrule
\multicolumn{4}{c}{Huffpost (2012 $\rightarrow$ 2017) (3 / 4 / 4)} \\ \midrule
\multicolumn{1}{c|}{Method} & $acc_{C}$ & $ acc_{\bar{C}_t} $ & H-score \\ \midrule
\multicolumn{1}{c|}{UDALM} & 32.85 {\footnotesize $\pm 1.05$} & 43.08 {\footnotesize $\pm 6.76$} & 37.08 {\footnotesize $\pm 2.82$} \\
\multicolumn{1}{c|}{AdSPT} & 35.74 {\footnotesize $\pm 3.33$} & 53.04 {\footnotesize $\pm 16.02$} & \underline{41.36} {\footnotesize $\pm 5.04$} \\
\midrule
\multicolumn{1}{c|}{OVANet} & 48.37 {\footnotesize $\pm 4.12$} & 35.03 {\footnotesize $\pm 3.60$} & 40.46 {\footnotesize $\pm 2.32$} \\
\multicolumn{1}{c|}{UniOT} & 31.57 {\footnotesize $\pm 1.92$} & 75.85 {\footnotesize $\pm 9.63$} & \textbf{44.36} {\footnotesize $\pm 1.27$} \\ \bottomrule
\end{tabular}

    }
    \caption{
        \label{table:domain_hard_results}
        Experimental results on MASSIVE and Huffpost (2017), which demonstrates high domain gap.
        The \textbf{best method} with the highest H-score is in bold, and the \underline{second-best method} is underlined.
    }
\end{table}

\subsection{Overview}
We conduct evaluations based on the clusters defined in Section \ref{sec:dataset_analysis} 
and analyze how the results vary depending on the adaptation complexity.
Figure \ref{fig:overal_results} presents an overview of the H-score results for the best-performing method from each CDA and UniDA approach: AdSPT representing CDA and UniOT representing UniDA.
Despite an outlier caused by unstable thresholding in CLINC-150, 
the overall trend demonstrates that AdSPT manifests comparable performance in less complex scenarios,
while UniOT exhibits superior performance towards challenging scenarios.
These trends align with the findings of other methods that are not depicted in the figure.

\subsection{Detailed Results}
% easy results
Table \ref{table:easy_results} demonstrates the results of relatively easy adaptation scenarios.
CDA methods demonstrate performance that is on par with, or even superior to, UniDA methods.
The results appear counter-intuitive, as CDA methods are designed without considering \textit{unknown} samples.
Specifically, UDALM outperforms all the UniDA methods in CLINC-150 and performs comparable or even better in Huffpost (2013).
AdSPT exhibits the best performance in Huffpost (2013). 
However, AdSPT suffers a significant performance drop in CLINC-150,
as we speculate this result is due to the inherent instability of the thresholding method.
The misguided threshold classifies the majority of the inputs as \textit{unknown}, which leads to a very high $acc_{\bar{C}_t}$,
but significantly reduces the $acc_{C}$.
This inconsistency also leads to a high variance of $acc_{\bar{C}_t}$ for all the CDA methods.

% moderate results
In the case of moderate shifts, no particular method decisively stands out, as presented by Table \ref{table:mid_results}.
In all cases, AdSPT and UniOT present the best performance with a marginal difference,
making it inconclusive to determine a superior approach.
Despite the relatively subpar performance, UDALM and OVANet also exhibit similar results.
Still, it is notable that CDA methods, which are not inherently optimized for UniDA settings, show comparable results.

% difficult category gap
The result of Amazon, in which the category gap is most prominent, is reported in Table \ref{table:category_hard_results}. 
UniDA methods exhibit substantially superior performance to CDA methods. 
In particular, while the difference in $acc_{C}$ is marginal, 
there exists a substantial gap of up to 55 points in $acc_{\bar{C}_t}$.
As the category gap intensifies, we observe the decline in the performance of CDA methods, 
which are fundamentally limited by the inability to handle \textit{unknown} inputs.

Finally, the results of Huffpost (2017) and MASSIVE, which exhibit a high domain gap, are reported in Table \ref{table:domain_hard_results}.
The result indicates that UniDA methods consistently display superior performance in most cases.
However, the divergence between the approaches is relatively small compared to Amazon.
UniOT demonstrates the best performance in all datasets, 
with OVANet's slightly lower performance.
AdSPT demonstrates marginally better performance than OVANet in Huffpost (2017), but the gap is marginal.
Even though CDA methods pose comparable performance,
UniDA methods demonstrate better overall performance.

\label{sec:analysis}

\subsection{Impact of Threshold Values}

\begin{figure*}[t]
    \centering
    \begin{subfigure}{0.23\textwidth}
        \includegraphics[width=\textwidth]{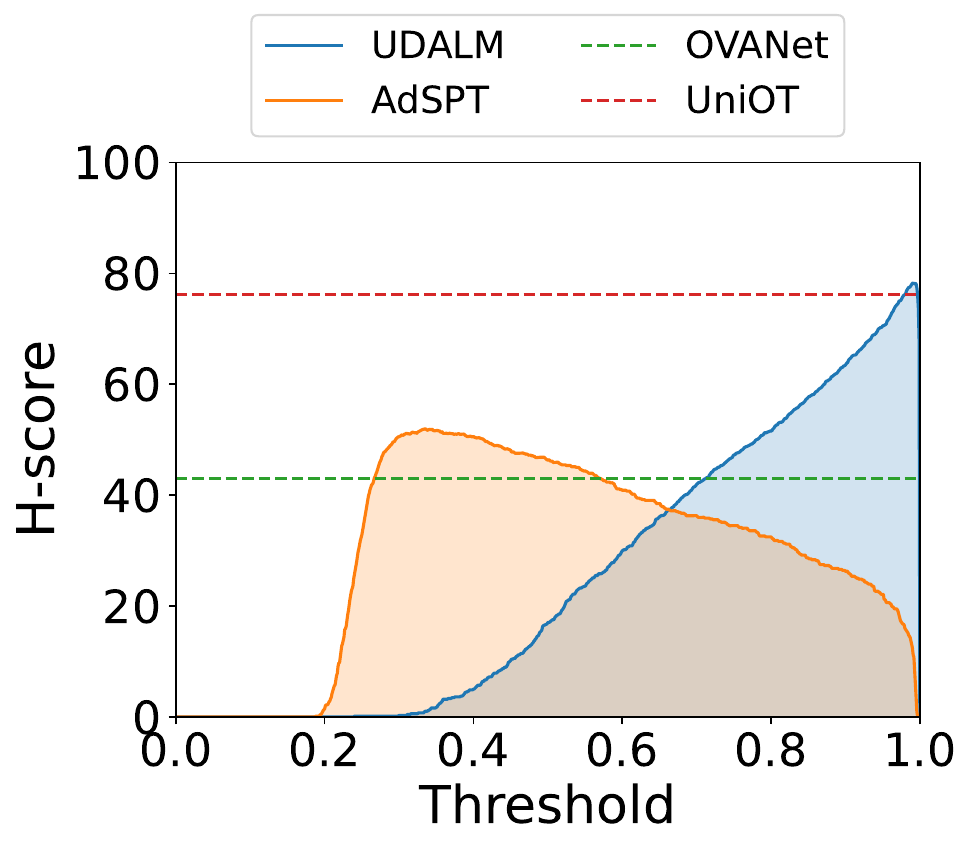}
        \caption{CLINC-150}
        \label{figure:different_thresholds_clinc}
    \end{subfigure}
    \hfill
    \begin{subfigure}{0.23\textwidth}
        \includegraphics[width=\textwidth]{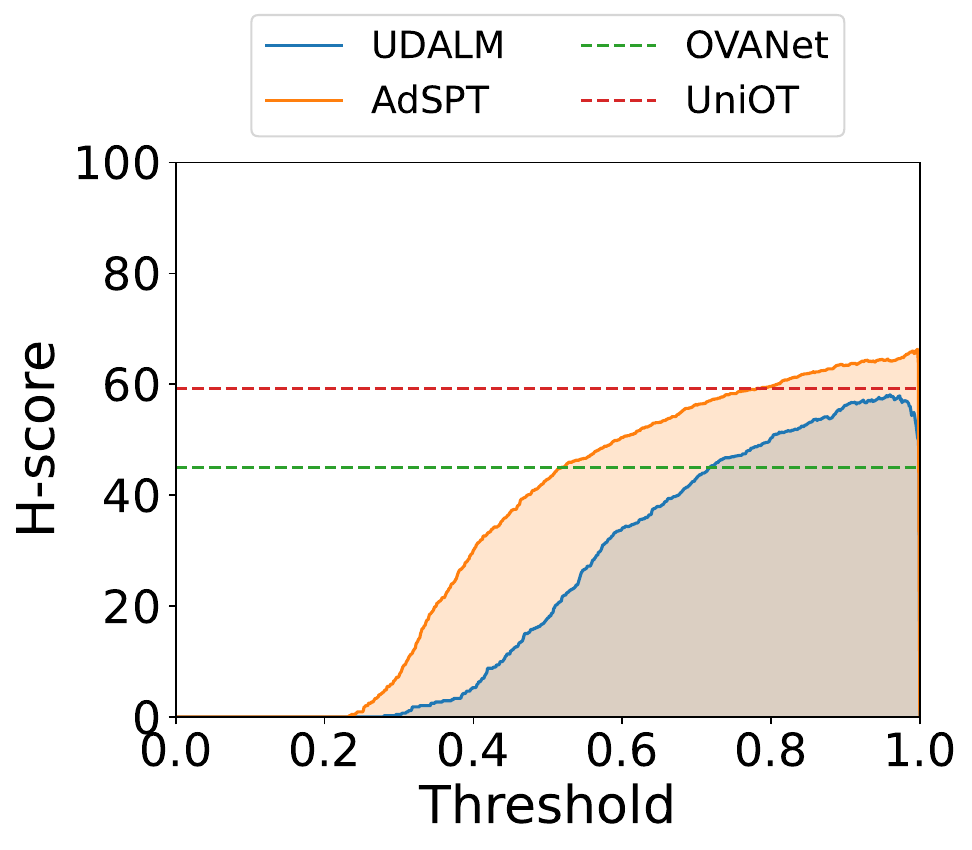}
        \caption{Huffpost (2012 $\rightarrow$ 2013)}
        \label{figure:different_thresholds_2013}
    \end{subfigure}
    \hfill
    \begin{subfigure}{0.23\textwidth}
        \includegraphics[width=\textwidth]{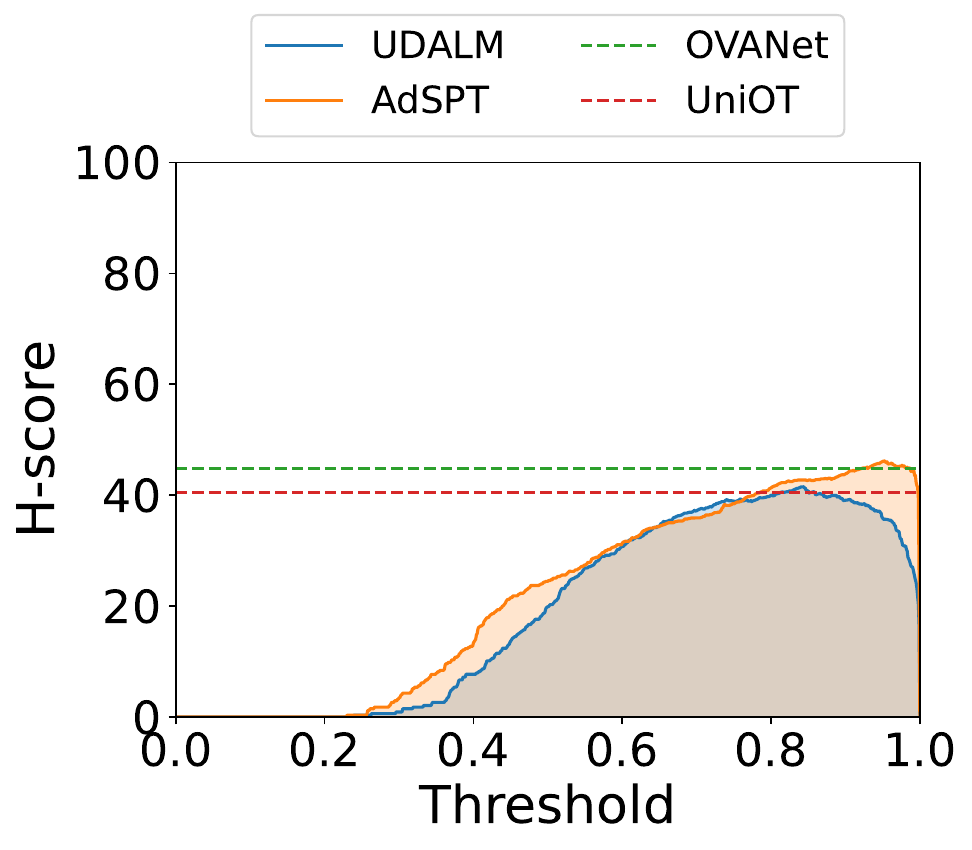}
        \caption{Huffpost (2012 $\rightarrow$ 2014)}
        \label{figure:different_thresholds_2014}
    \end{subfigure}
    \hfill
    \begin{subfigure}{0.23\textwidth}
        \includegraphics[width=\textwidth]{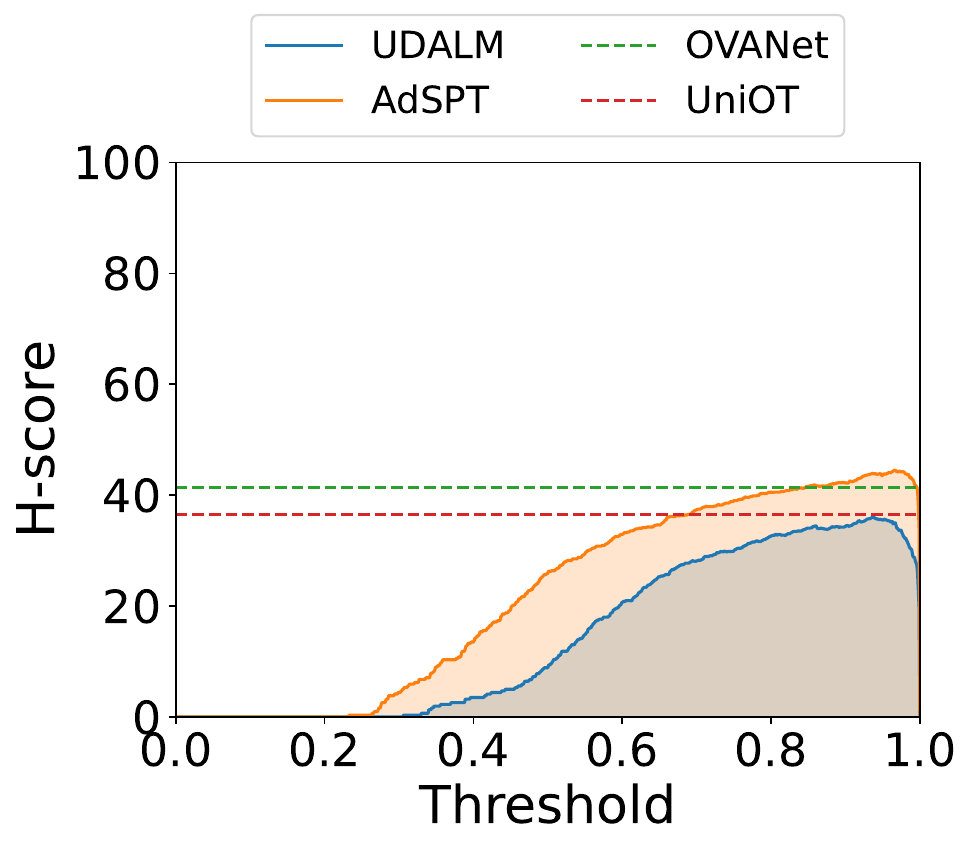}
        \caption{Huffpost (2012 $\rightarrow$ 2015)}
        \label{figure:different_thresholds_2015}
    \end{subfigure}
    \hfill
    \begin{subfigure}{0.23\textwidth}
        \includegraphics[width=\textwidth]{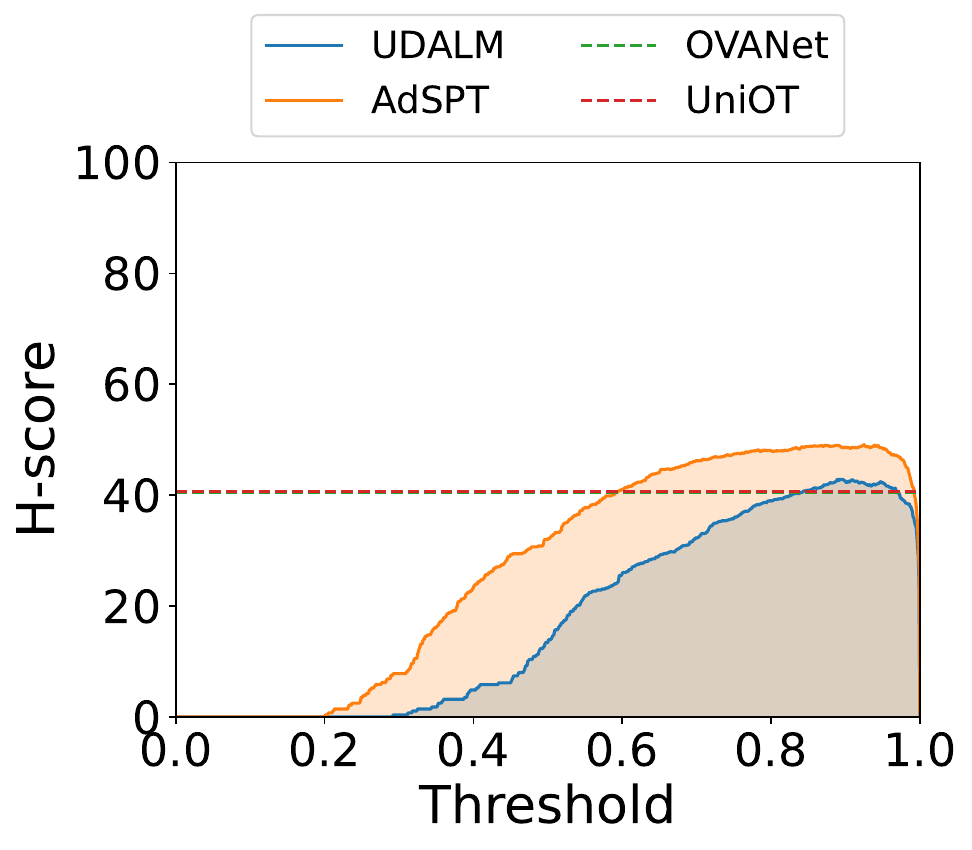}
        \caption{Huffpost (2012 $\rightarrow$ 2016)}
        \label{figure:different_thresholds_2016}
    \end{subfigure}
    \hfill
    \begin{subfigure}{0.23\textwidth}
        \includegraphics[width=\textwidth]{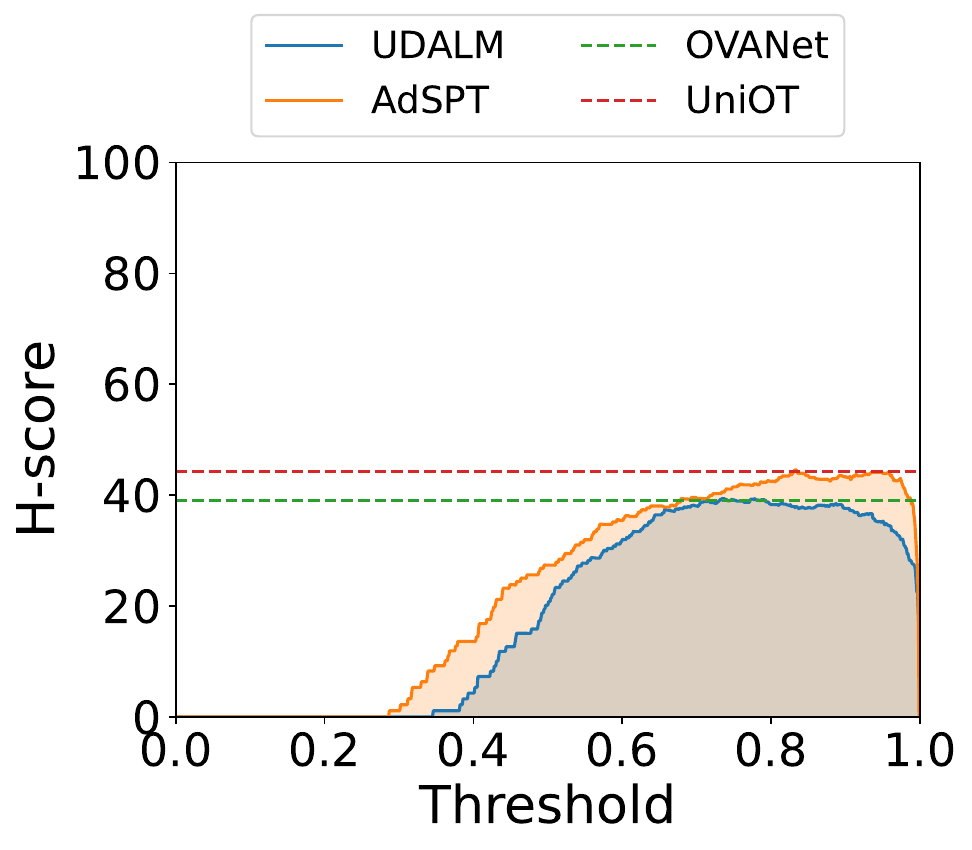}
        \caption{Huffpost (2012 $\rightarrow$ 2017)}
        \label{figure:different_thresholds_2017}
    \end{subfigure}
    \hfill
    \begin{subfigure}{0.23\textwidth}
        \includegraphics[width=\textwidth]{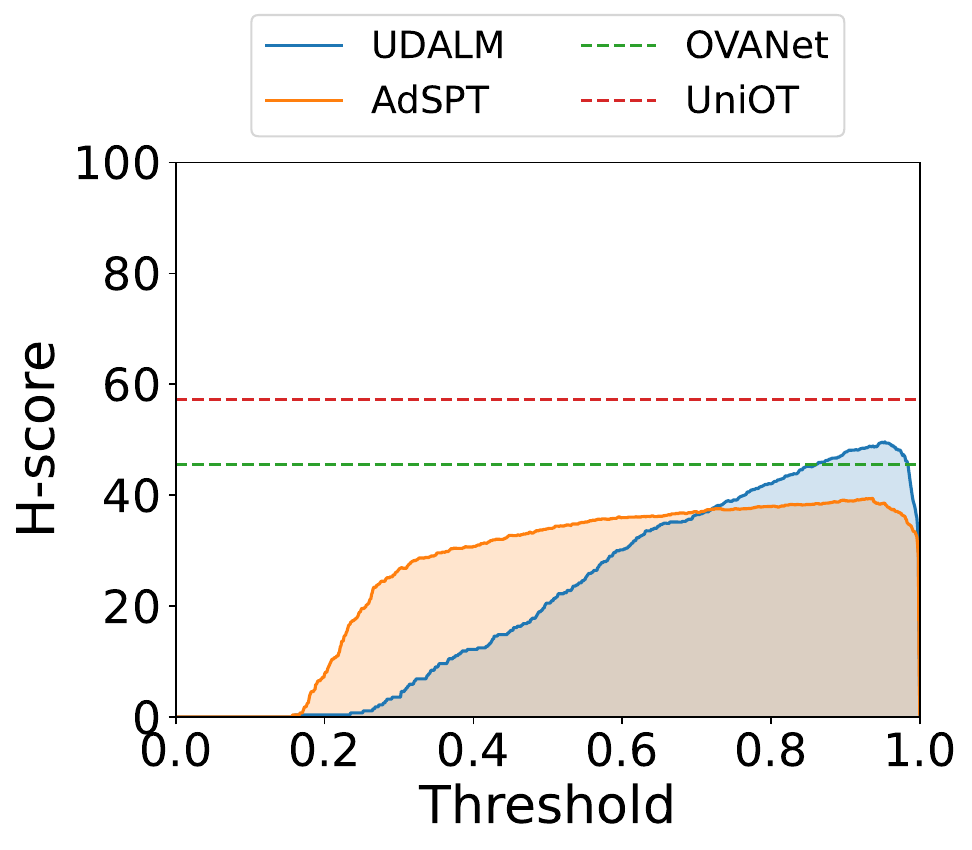}
        \caption{MASSIVE}
        \label{figure:different_thresholds_massive}
    \end{subfigure}
    \hfill
    \begin{subfigure}{0.23\textwidth}
        \includegraphics[width=\textwidth]{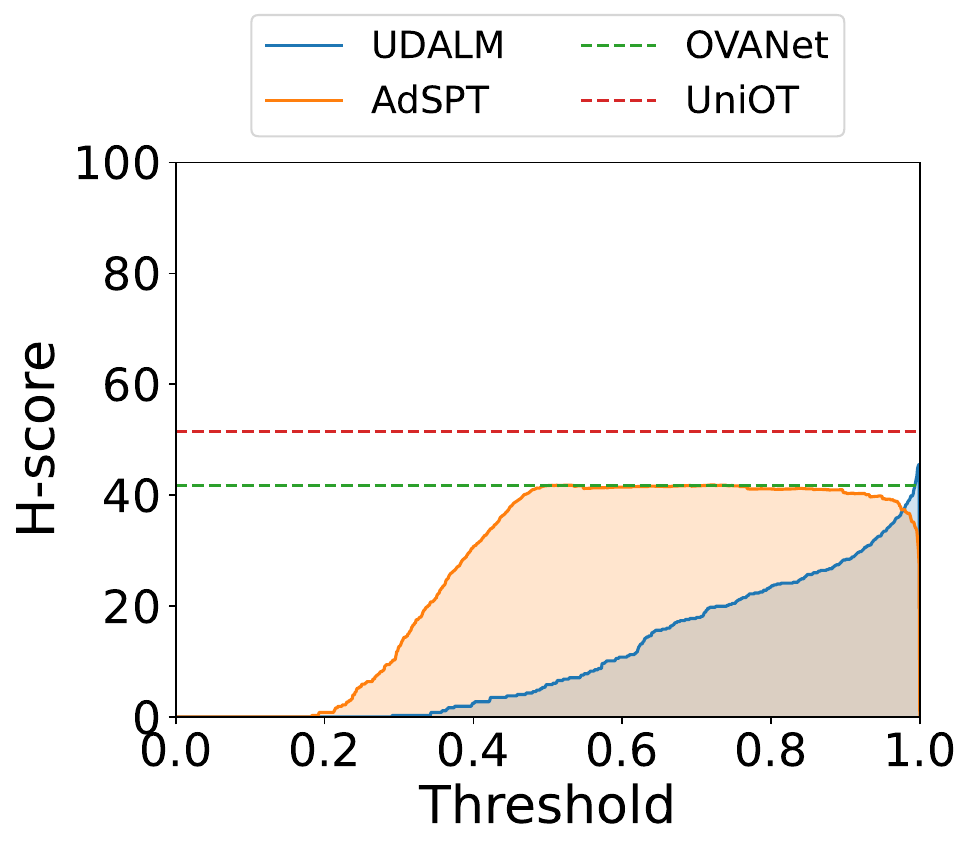}
        \caption{Amazon}
        \label{figure:different_thresholds_amazon}
    \end{subfigure}
            
    \caption{
        H-score performance with different threshold values. 
        Results of UniDA methods are visualized as a horizontal line for comparison.
    }
    \label{figure:different_thresholds}
\end{figure*}

% \subsection{Various Threshold Values}
\label{sec:different_thresholds}

The selection of the threshold value considerably influences the performance of CDA methods.
In order to probe the impact of the threshold values on the performance, 
we carry out an analysis whereby different threshold values are applied to measure the performance of the methods.
    
The results are demonstrated in Figure \ref{figure:different_thresholds}.
In cases of low or moderate adaptation complexity, such as CLINC-150 and Huffpost (2013, 2014, 2015, 2016), 
CDA methods demonstrate the potential to outperform UniDA methods when provided an appropriate threshold.
However, as the adaptation complexity intensifies, such as Huffpost (2017), MASSIVE, and Amazon, 
UniDA methods outperform CDA methods regardless of the selected threshold.
These observations align seamlessly with the findings from Section \ref{sec:results} 
that underscore the proficiency of UniDA methods in managing challenging adaptation scenarios.
Additionally, it should be noted that determining the optimal threshold is particularly challenging in the absence of supervision from the target domain.
Therefore, the best performance should be considered upper-bound of the CDA methods.

\label{sec:related_work}

%%%%%%%%%%%%%%%%%%%%%%%%%%%%%%%%%%%%%%%%%%%%%%%%%%%%%%%%%%%%%%%%%%%%%%%%%%%%%%%%%%%%%%%%%%%%%%%%%%%%%%%%%%%%%

\section{Related Work}

\subsection{Domain Adaptation}
% The main task of DA is to adapt from the source domain to the target domain of a shifted distribution. 
% The most common assumption in the DA literature is the closed-set assumption, where the source and the target domain share the same class set. 
% Closed-set domain adaption (CDA) studies mainly focus on learning domain-invariant features \citep{blitzer-etal-2006-domain,10.1145/1772690.1772767,ben2020perl,10.5555/3045118.3045244,du2020adversarial} for adaptation.
% Recent methods utilized pre-trained language model (PLMs) \citep{devlin-etal-2019-bert,liu2020roberta} by
% applying masked language modeling \citep{Karouzos2021UDALMUD} or soft-prompt with adversarial training \citep{wu-shi-2022-adversarial}.
% Nevertheless, the closed-set assumption have a fundamental limitation as they may be vulnerable when confronting data from an unknown class.
The studies in the field of DA in NLP primarily assumes a closed-set environment, which the source and the target domain share the same label space.
CDA research predominantly concentrated on learning domain invariant features \citep{blitzer-etal-2006-domain,10.1145/1772690.1772767,ben2020perl,10.5555/3045118.3045244,du2020adversarial} for effective adaptation. With the advent of pre-trained language models (PLMs), CDA methods have evolved to effectively leverage the capabilities of PLMs. Techniques such as masked language modeling \citep{Karouzos2021UDALMUD} or soft-prompt with adversarial training \citep{wu-shi-2022-adversarial} have shown promising results. However, the closed-set assumption has a fundamental drawback as it may leave the models vulnerable when exposed to data from an unknown class.

To mitigate such issue, a new line of work named UniDA \citep{UDA_2019_CVPR} was proposed which assumes no prior knowledge about the target domain.
\citet{UDA_2019_CVPR} quantifies sample-level transferability by using of uncertainty and domain similarity. 
Following the work, \citet{Fu2020LearningTD} calibrates multiple uncertainty measures to handle such an issue.
\citet{saito2021ovanet} apply a one-vs-all classifier to minimize inter-class distance and classify unknown classes.
More recently, \citet{chang2022unified} applied Optimal Transport and further expanded the task to discovering private classes.
Other recent works focus on utilizing mutually nearest neighbor samples \citep{Chen_Du_Lou_He_Bai_Deng_2022,9878972} 
or leveraging source prototypes with target samples \citep{Chen_Lou_He_Bai_Deng_2022,kundu2022subsidiary}.
Despite the practicality of UniDA, its application in the NLP domain has barely explored.

%%%%%%%%%%%%%%%%%%%%%%%%%%%%%%%%%%%%%%%%%%%%%%%%%%%%%%%%%%%%%%%%%%%%%%%%%%%%%%%%%%%%%%%%%%%%%%%%%%%%%%%%%%%%%

\subsection{Out-of-Distribution Detection}
The early exploration of OOD detection focused on training supervised detectors \citep{dhamija2018reducing,lee2018training, jiang2018trust}. However, since obtaining labeled OOD samples is impractical, recent OOD detection research has shifted towards unsupervised methods, such as generating pseudo-OOD data \citep{chen-yu-2021-gold,zheng2020out}, utilizing self-supervised learning \citep{Moon_Mo_Lee_Lee_Shin_2021,manolache-etal-2021-date,9414930,zeng-etal-2021-modeling,zhan-etal-2021-scope, cho-etal-2022-enhancing}, and measuring uncertainty through scoring functions for input instances \citep{hendrycks2017a,NEURIPS2018_abdeb6f5,NEURIPS2020_f5496252,NEURIPS2020_8965f766}. While these methods have shown effectiveness, OOD detection is limited in that it does not offer opportunities for adaptation.

\label{sec:conclusion}
\section{Conclusion and Future Work}
In this study, we present a testbed for evaluating UniDA in the field of NLP.
The testbed is designed to exhibit various levels of domain and category gaps through different datasets. 
Two novel metrics, PDR and DDS, were proposed which can measure the degree of domain and category gap, respectively.
We assessed UniDA methods and the heuristic combination of CDA and OOD detection in our proposed testbed.
Experimental results show that UniDA methods, initially designed for the vision domain, can be effectively transferred to NLP.
Additionally, CDA methods, which are not fully optimized in UniDA scenario, produce comparable results in certain circumstances.

Recent trends in NLP focus on Large Language Models (LLMs) of their significant generalization abilities.
However, the robustness of LLMs from the perspective of UniDA remains uncertain. 
As part of our future work, we assess the performance and the capabilities of LLMs from a UniDA viewpoint.

\label{sec:limitations}
\section*{Limitations}

\paragraph{Limited coverage of the evaluated model sizes} 
The evaluation was conducted only with models of limited size. 
Moreover, there is a lack of zero-shot and few-shot evaluations for large language models (LLMs) that have recently emerged with remarkable generalization capabilities. 
The evaluation of LLMs is currently being considered as a top priority for our future work, and based on preliminary experiments, the results were somewhat unsatisfactory compared to small models with basic tuning for classification performance. 
In this regard, recent research that evaluated LLMs for classification problems (such as GLUE) also reported that the performance is not yet comparable to task-specifically tuned models. 
Considering the limitations of LLMs in terms of their massive resource usage and the fact that tuning small models still outperforms them in a task-specific manner, 
the findings from this study are still considered highly valuable in the NLP community.

\paragraph{Limited scope of the tasks} 
Our proposed testbed is restricted to text classification tasks only.
The majority of existing research on DA and OOD also focuses on classification. 
This selective task preference is primarily due to the challenge of defining concepts such as domain shifts and category shifts in generative tasks.
However, in light of recent advancements in the generative capabilities of models, 
handling distributional shifts in generative tasks is indubitably an essential problem that needs to be addressed in our future work.

\section*{Acknowledgement}
This work was mainly supported by SNU-NAVER Hyperscale AI Center and partly supported by Institute of Information \& communications Technology Planning \& Evaluation (IITP) grant funded by the Korea government (MSIT) 
[No.2020-0-01373, Artificial Intelligence Graduate School Program (Hanyang University), 
NO.2021-0-02068, Artificial Intelligence Innovation Hub (Artificial Intelligence Institute, Seoul National University)]
and Korea Evaluation Institute of Industrial Technology (KEIT) grant funded by the Korea government (MOTIE).

% \section*{Ethics Statement}
% Scientific work published at EMNLP 2023 must comply with the \href{https://www.aclweb.org/portal/content/acl-code-ethics}{ACL Ethics Policy}. We encourage all authors to include an explicit ethics statement on the broader impact of the work, or other ethical considerations after the conclusion but before the references. The ethics statement will not count toward the page limit (8 pages for long, 4 pages for short papers).

\bibliography{anthology,custom}

\begin{thebibliography}{51}
\expandafter\ifx\csname natexlab\endcsname\relax\def\natexlab#1{#1}\fi

\bibitem[{Aggarwal(2017)}]{Aggarwal2017}
Charu~C. Aggarwal. 2017.
\newblock \href {https://doi.org/10.1007/978-3-319-47578-3_1} {\emph{An Introduction to Outlier Analysis}}, pages 1--34. Springer International Publishing, Cham.

\bibitem[{Ben-David et~al.(2020)Ben-David, Rabinovitz, and Reichart}]{ben2020perl}
Eyal Ben-David, Carmel Rabinovitz, and Roi Reichart. 2020.
\newblock Perl: Pivot-based domain adaptation for pre-trained deep contextualized embedding models.
\newblock \emph{Transactions of the Association for Computational Linguistics}, 8:504--521.

\bibitem[{Blitzer et~al.(2006)Blitzer, McDonald, and Pereira}]{blitzer-etal-2006-domain}
John Blitzer, Ryan McDonald, and Fernando Pereira. 2006.
\newblock \href {https://aclanthology.org/W06-1615} {Domain adaptation with structural correspondence learning}.
\newblock In \emph{Proceedings of the 2006 Conference on Empirical Methods in Natural Language Processing}, pages 120--128, Sydney, Australia. Association for Computational Linguistics.

\bibitem[{Brown et~al.(2020)Brown, Mann, Ryder, Subbiah, Kaplan, Dhariwal, Neelakantan, Shyam, Sastry, Askell, Agarwal, Herbert-Voss, Krueger, Henighan, Child, Ramesh, Ziegler, Wu, Winter, Hesse, Chen, Sigler, Litwin, Gray, Chess, Clark, Berner, McCandlish, Radford, Sutskever, and Amodei}]{NEURIPS2020_1457c0d6}
Tom Brown, Benjamin Mann, Nick Ryder, Melanie Subbiah, Jared~D Kaplan, Prafulla Dhariwal, Arvind Neelakantan, Pranav Shyam, Girish Sastry, Amanda Askell, Sandhini Agarwal, Ariel Herbert-Voss, Gretchen Krueger, Tom Henighan, Rewon Child, Aditya Ramesh, Daniel Ziegler, Jeffrey Wu, Clemens Winter, Chris Hesse, Mark Chen, Eric Sigler, Mateusz Litwin, Scott Gray, Benjamin Chess, Jack Clark, Christopher Berner, Sam McCandlish, Alec Radford, Ilya Sutskever, and Dario Amodei. 2020.
\newblock \href {https://proceedings.neurips.cc/paper_files/paper/2020/file/1457c0d6bfcb4967418bfb8ac142f64a-Paper.pdf} {Language models are few-shot learners}.
\newblock In \emph{Advances in Neural Information Processing Systems}, volume~33, pages 1877--1901. Curran Associates, Inc.

\bibitem[{Chang et~al.(2022)Chang, Shi, Tuan, and Wang}]{chang2022unified}
Wanxing Chang, Ye~Shi, Hoang~Duong Tuan, and Jingya Wang. 2022.
\newblock \href {https://openreview.net/forum?id=RTan64GlCLV} {Unified optimal transport framework for universal domain adaptation}.
\newblock In \emph{Advances in Neural Information Processing Systems}.

\bibitem[{Chen and Yu(2021)}]{chen-yu-2021-gold}
Derek Chen and Zhou Yu. 2021.
\newblock \href {https://doi.org/10.18653/v1/2021.emnlp-main.35} {{GOLD}: Improving out-of-scope detection in dialogues using data augmentation}.
\newblock In \emph{Proceedings of the 2021 Conference on Empirical Methods in Natural Language Processing}, pages 429--442, Online and Punta Cana, Dominican Republic. Association for Computational Linguistics.

\bibitem[{Chen et~al.(2022{\natexlab{a}})Chen, Du, Lou, He, Bai, and Deng}]{Chen_Du_Lou_He_Bai_Deng_2022}
Liang Chen, Qianjin Du, Yihang Lou, Jianzhong He, Tao Bai, and Minghua Deng. 2022{\natexlab{a}}.
\newblock \href {https://doi.org/10.1609/aaai.v36i6.20574} {Mutual nearest neighbor contrast and hybrid prototype self-training for universal domain adaptation}.
\newblock \emph{Proceedings of the AAAI Conference on Artificial Intelligence}, 36(6):6248--6257.

\bibitem[{Chen et~al.(2022{\natexlab{b}})Chen, Lou, He, Bai, and Deng}]{Chen_Lou_He_Bai_Deng_2022}
Liang Chen, Yihang Lou, Jianzhong He, Tao Bai, and Minghua Deng. 2022{\natexlab{b}}.
\newblock \href {https://doi.org/10.1609/aaai.v36i6.20575} {Evidential neighborhood contrastive learning for universal domain adaptation}.
\newblock \emph{Proceedings of the AAAI Conference on Artificial Intelligence}, 36(6):6258--6267.

\bibitem[{Chen et~al.(2022{\natexlab{c}})Chen, Lou, He, Bai, and Deng}]{9878972}
Liang Chen, Yihang Lou, Jianzhong He, Tao Bai, and Minghua Deng. 2022{\natexlab{c}}.
\newblock \href {https://doi.org/10.1109/CVPR52688.2022.01566} {Geometric anchor correspondence mining with uncertainty modeling for universal domain adaptation}.
\newblock In \emph{2022 IEEE/CVF Conference on Computer Vision and Pattern Recognition (CVPR)}, pages 16113--16122.

\bibitem[{Cho et~al.(2022)Cho, Park, Kang, Yoo, Kim, and Lee}]{cho-etal-2022-enhancing}
Hyunsoo Cho, Choonghyun Park, Jaewook Kang, Kang~Min Yoo, Taeuk Kim, and Sang-goo Lee. 2022.
\newblock \href {https://aclanthology.org/2022.findings-emnlp.55} {Enhancing out-of-distribution detection in natural language understanding via implicit layer ensemble}.
\newblock In \emph{Findings of the Association for Computational Linguistics: EMNLP 2022}, pages 783--798, Abu Dhabi, United Arab Emirates. Association for Computational Linguistics.

\bibitem[{Cho et~al.(2021)Cho, Seol, and Lee}]{ijcai2021p198}
Hyunsoo Cho, Jinseok Seol, and Sang-goo Lee. 2021.
\newblock \href {https://doi.org/10.24963/ijcai.2021/198} {Masked contrastive learning for anomaly detection}.
\newblock In \emph{Proceedings of the Thirtieth International Joint Conference on Artificial Intelligence, {IJCAI-21}}, pages 1434--1441. International Joint Conferences on Artificial Intelligence Organization.
\newblock Main Track.

\bibitem[{Chowdhery et~al.(2022)Chowdhery, Narang, Devlin, Bosma, Mishra, Roberts, Barham, Chung, Sutton, Gehrmann, Schuh, Shi, Tsvyashchenko, Maynez, Rao, Barnes, Tay, Shazeer, Prabhakaran, Reif, Du, Hutchinson, Pope, Bradbury, Austin, Isard, Gur-Ari, Yin, Duke, Levskaya, Ghemawat, Dev, Michalewski, Garcia, Misra, Robinson, Fedus, Zhou, Ippolito, Luan, Lim, Zoph, Spiridonov, Sepassi, Dohan, Agrawal, Omernick, Dai, Pillai, Pellat, Lewkowycz, Moreira, Child, Polozov, Lee, Zhou, Wang, Saeta, Diaz, Firat, Catasta, Wei, Meier-Hellstern, Eck, Dean, Petrov, and Fiedel}]{chowdhery2022palm}
Aakanksha Chowdhery, Sharan Narang, Jacob Devlin, Maarten Bosma, Gaurav Mishra, Adam Roberts, Paul Barham, Hyung~Won Chung, Charles Sutton, Sebastian Gehrmann, Parker Schuh, Kensen Shi, Sasha Tsvyashchenko, Joshua Maynez, Abhishek Rao, Parker Barnes, Yi~Tay, Noam Shazeer, Vinodkumar Prabhakaran, Emily Reif, Nan Du, Ben Hutchinson, Reiner Pope, James Bradbury, Jacob Austin, Michael Isard, Guy Gur-Ari, Pengcheng Yin, Toju Duke, Anselm Levskaya, Sanjay Ghemawat, Sunipa Dev, Henryk Michalewski, Xavier Garcia, Vedant Misra, Kevin Robinson, Liam Fedus, Denny Zhou, Daphne Ippolito, David Luan, Hyeontaek Lim, Barret Zoph, Alexander Spiridonov, Ryan Sepassi, David Dohan, Shivani Agrawal, Mark Omernick, Andrew~M. Dai, Thanumalayan~Sankaranarayana Pillai, Marie Pellat, Aitor Lewkowycz, Erica Moreira, Rewon Child, Oleksandr Polozov, Katherine Lee, Zongwei Zhou, Xuezhi Wang, Brennan Saeta, Mark Diaz, Orhan Firat, Michele Catasta, Jason Wei, Kathy Meier-Hellstern, Douglas Eck, Jeff Dean, Slav Petrov, and Noah Fiedel. 2022.
\newblock \href {http://arxiv.org/abs/2204.02311} {Palm: Scaling language modeling with pathways}.

\bibitem[{Devlin et~al.(2019)Devlin, Chang, Lee, and Toutanova}]{devlin-etal-2019-bert}
Jacob Devlin, Ming-Wei Chang, Kenton Lee, and Kristina Toutanova. 2019.
\newblock \href {https://doi.org/10.18653/v1/N19-1423} {{BERT}: Pre-training of deep bidirectional transformers for language understanding}.
\newblock In \emph{Proceedings of the 2019 Conference of the North {A}merican Chapter of the Association for Computational Linguistics: Human Language Technologies, Volume 1 (Long and Short Papers)}, pages 4171--4186, Minneapolis, Minnesota. Association for Computational Linguistics.

\bibitem[{Dhamija et~al.(2018)Dhamija, G{\"u}nther, and Boult}]{dhamija2018reducing}
Akshay~Raj Dhamija, Manuel G{\"u}nther, and Terrance Boult. 2018.
\newblock Reducing network agnostophobia.
\newblock \emph{Advances in Neural Information Processing Systems}, 31.

\bibitem[{Du et~al.(2020)Du, Sun, Wang, Qi, and Liao}]{du2020adversarial}
Chunning Du, Haifeng Sun, Jingyu Wang, Qi~Qi, and Jianxin Liao. 2020.
\newblock Adversarial and domain-aware bert for cross-domain sentiment analysis.
\newblock In \emph{Proceedings of the 58th annual meeting of the Association for Computational Linguistics}, pages 4019--4028.

\bibitem[{FitzGerald et~al.(2022)FitzGerald, Hench, Peris, Mackie, Rottmann, Sanchez, Nash, Urbach, Kakarala, Singh, Ranganath, Crist, Britan, Leeuwis, Tur, and Natarajan}]{fitzgerald2022massive}
Jack FitzGerald, Christopher Hench, Charith Peris, Scott Mackie, Kay Rottmann, Ana Sanchez, Aaron Nash, Liam Urbach, Vishesh Kakarala, Richa Singh, Swetha Ranganath, Laurie Crist, Misha Britan, Wouter Leeuwis, Gokhan Tur, and Prem Natarajan. 2022.
\newblock \href {http://arxiv.org/abs/2204.08582} {Massive: A 1m-example multilingual natural language understanding dataset with 51 typologically-diverse languages}.

\bibitem[{Fu et~al.(2020)Fu, Cao, Long, and Wang}]{Fu2020LearningTD}
Bo~Fu, Zhangjie Cao, Mingsheng Long, and Jianmin Wang. 2020.
\newblock Learning to detect open classes for universal domain adaptation.
\newblock In \emph{European Conference on Computer Vision}.

\bibitem[{Ganin and Lempitsky(2015)}]{10.5555/3045118.3045244}
Yaroslav Ganin and Victor Lempitsky. 2015.
\newblock Unsupervised domain adaptation by backpropagation.
\newblock In \emph{Proceedings of the 32nd International Conference on International Conference on Machine Learning - Volume 37}, ICML'15, page 1180–1189. JMLR.org.

\bibitem[{Ganin et~al.(2016{\natexlab{a}})Ganin, Ustinova, Ajakan, Germain, Larochelle, Laviolette, Marchand, and Lempitsky}]{Ganin2016DomainAdversarialTO}
Yaroslav Ganin, E.~Ustinova, Hana Ajakan, Pascal Germain, H.~Larochelle, François Laviolette, Mario Marchand, and Victor~S. Lempitsky. 2016{\natexlab{a}}.
\newblock Domain-adversarial training of neural networks.
\newblock In \emph{Journal of machine learning research}.

\bibitem[{Ganin et~al.(2016{\natexlab{b}})Ganin, Ustinova, Ajakan, Germain, Larochelle, Laviolette, Marchand, and Lempitsky}]{ganin2016domain}
Yaroslav Ganin, Evgeniya Ustinova, Hana Ajakan, Pascal Germain, Hugo Larochelle, Fran{\c{c}}ois Laviolette, Mario Marchand, and Victor Lempitsky. 2016{\natexlab{b}}.
\newblock Domain-adversarial training of neural networks.
\newblock \emph{The journal of machine learning research}, 17(1):2096--2030.

\bibitem[{Hendrycks and Gimpel(2017)}]{hendrycks2017a}
Dan Hendrycks and Kevin Gimpel. 2017.
\newblock \href {https://openreview.net/forum?id=Hkg4TI9xl} {A baseline for detecting misclassified and out-of-distribution examples in neural networks}.
\newblock In \emph{International Conference on Learning Representations}.

\bibitem[{Hendrycks et~al.(2020)Hendrycks, Liu, Wallace, Dziedzic, Krishnan, and Song}]{hendrycks-etal-2020-pretrained}
Dan Hendrycks, Xiaoyuan Liu, Eric Wallace, Adam Dziedzic, Rishabh Krishnan, and Dawn Song. 2020.
\newblock \href {https://doi.org/10.18653/v1/2020.acl-main.244} {Pretrained transformers improve out-of-distribution robustness}.
\newblock In \emph{Proceedings of the 58th Annual Meeting of the Association for Computational Linguistics}, pages 2744--2751, Online. Association for Computational Linguistics.

\bibitem[{Hendrycks et~al.(2019)Hendrycks, Mazeika, and Dietterich}]{hendrycks2018deep}
Dan Hendrycks, Mantas Mazeika, and Thomas Dietterich. 2019.
\newblock \href {https://openreview.net/forum?id=HyxCxhRcY7} {Deep anomaly detection with outlier exposure}.
\newblock In \emph{International Conference on Learning Representations}.

\bibitem[{Jiang et~al.(2018)Jiang, Kim, Guan, and Gupta}]{jiang2018trust}
Heinrich Jiang, Been Kim, Melody Guan, and Maya Gupta. 2018.
\newblock To trust or not to trust a classifier.
\newblock \emph{Advances in neural information processing systems}, 31.

\bibitem[{Karouzos et~al.(2021)Karouzos, Paraskevopoulos, and Potamianos}]{Karouzos2021UDALMUD}
Constantinos~F. Karouzos, Georgios Paraskevopoulos, and Alexandros Potamianos. 2021.
\newblock Udalm: Unsupervised domain adaptation through language modeling.
\newblock \emph{ArXiv}, abs/2104.07078.

\bibitem[{Keung et~al.(2020)Keung, Lu, Szarvas, and Smith}]{keung-etal-2020-multilingual}
Phillip Keung, Yichao Lu, Gy{\"o}rgy Szarvas, and Noah~A. Smith. 2020.
\newblock \href {https://doi.org/10.18653/v1/2020.emnlp-main.369} {The multilingual {A}mazon reviews corpus}.
\newblock In \emph{Proceedings of the 2020 Conference on Empirical Methods in Natural Language Processing (EMNLP)}, pages 4563--4568, Online. Association for Computational Linguistics.

\bibitem[{Kundu et~al.(2022)Kundu, Bhambri, Kulkarni, Sarkar, Jampani, and Babu}]{kundu2022subsidiary}
Jogendra~Nath Kundu, Suvaansh Bhambri, Akshay Kulkarni, Hiran Sarkar, Varun Jampani, and R.~Venkatesh Babu. 2022.
\newblock Subsidiary prototype alignment for universal domain adaptation.
\newblock In \emph{Advances in Neural Information Processing Systems (NeurIPS)}.

\bibitem[{Larson et~al.(2019)Larson, Mahendran, Peper, Clarke, Lee, Hill, Kummerfeld, Leach, Laurenzano, Tang, and Mars}]{larson-etal-2019-evaluation}
Stefan Larson, Anish Mahendran, Joseph~J. Peper, Christopher Clarke, Andrew Lee, Parker Hill, Jonathan~K. Kummerfeld, Kevin Leach, Michael~A. Laurenzano, Lingjia Tang, and Jason Mars. 2019.
\newblock \href {https://doi.org/10.18653/v1/D19-1131} {An evaluation dataset for intent classification and out-of-scope prediction}.
\newblock In \emph{Proceedings of the 2019 Conference on Empirical Methods in Natural Language Processing and the 9th International Joint Conference on Natural Language Processing (EMNLP-IJCNLP)}, pages 1311--1316, Hong Kong, China. Association for Computational Linguistics.

\bibitem[{Lee et~al.(2018{\natexlab{a}})Lee, Lee, Lee, and Shin}]{lee2018training}
Kimin Lee, Honglak Lee, Kibok Lee, and Jinwoo Shin. 2018{\natexlab{a}}.
\newblock \href {https://openreview.net/forum?id=ryiAv2xAZ} {Training confidence-calibrated classifiers for detecting out-of-distribution samples}.
\newblock In \emph{International Conference on Learning Representations}.

\bibitem[{Lee et~al.(2018{\natexlab{b}})Lee, Lee, Lee, and Shin}]{NEURIPS2018_abdeb6f5}
Kimin Lee, Kibok Lee, Honglak Lee, and Jinwoo Shin. 2018{\natexlab{b}}.
\newblock \href {https://proceedings.neurips.cc/paper_files/paper/2018/file/abdeb6f575ac5c6676b747bca8d09cc2-Paper.pdf} {A simple unified framework for detecting out-of-distribution samples and adversarial attacks}.
\newblock In \emph{Advances in Neural Information Processing Systems}, volume~31. Curran Associates, Inc.

\bibitem[{Li et~al.(2021)Li, Chen, Zhang, Dong, and Keutzer}]{9414930}
Tian Li, Xiang Chen, Shanghang Zhang, Zhen Dong, and Kurt Keutzer. 2021.
\newblock \href {https://doi.org/10.1109/ICASSP39728.2021.9414930} {Cross-domain sentiment classification with contrastive learning and mutual information maximization}.
\newblock In \emph{ICASSP 2021 - 2021 IEEE International Conference on Acoustics, Speech and Signal Processing (ICASSP)}, pages 8203--8207.

\bibitem[{Liu et~al.(2020)Liu, Wang, Owens, and Li}]{NEURIPS2020_f5496252}
Weitang Liu, Xiaoyun Wang, John Owens, and Yixuan Li. 2020.
\newblock \href {https://proceedings.neurips.cc/paper/2020/file/f5496252609c43eb8a3d147ab9b9c006-Paper.pdf} {Energy-based out-of-distribution detection}.
\newblock In \emph{Advances in Neural Information Processing Systems}, volume~33, pages 21464--21475. Curran Associates, Inc.

\bibitem[{Loshchilov and Hutter(2019)}]{Loshchilov2019DecoupledWD}
Ilya Loshchilov and Frank Hutter. 2019.
\newblock Decoupled weight decay regularization.
\newblock In \emph{International Conference on Learning Representations}.

\bibitem[{Manolache et~al.(2021)Manolache, Brad, and Burceanu}]{manolache-etal-2021-date}
Andrei Manolache, Florin Brad, and Elena Burceanu. 2021.
\newblock \href {https://doi.org/10.18653/v1/2021.naacl-main.25} {{DATE}: Detecting anomalies in text via self-supervision of transformers}.
\newblock In \emph{Proceedings of the 2021 Conference of the North American Chapter of the Association for Computational Linguistics: Human Language Technologies}, pages 267--277, Online. Association for Computational Linguistics.

\bibitem[{Miller et~al.(2020)Miller, Krauth, Recht, and Schmidt}]{pmlr-v119-miller20a}
John Miller, Karl Krauth, Benjamin Recht, and Ludwig Schmidt. 2020.
\newblock \href {https://proceedings.mlr.press/v119/miller20a.html} {The effect of natural distribution shift on question answering models}.
\newblock In \emph{Proceedings of the 37th International Conference on Machine Learning}, volume 119 of \emph{Proceedings of Machine Learning Research}, pages 6905--6916. PMLR.

\bibitem[{Misra(2022)}]{misra2022news}
Rishabh Misra. 2022.
\newblock News category dataset.
\newblock \emph{arXiv preprint arXiv:2209.11429}.

\bibitem[{Moon et~al.(2021)Moon, Mo, Lee, Lee, and Shin}]{Moon_Mo_Lee_Lee_Shin_2021}
Seung~Jun Moon, Sangwoo Mo, Kimin Lee, Jaeho Lee, and Jinwoo Shin. 2021.
\newblock \href {https://doi.org/10.1609/aaai.v35i15.17601} {Masker: Masked keyword regularization for reliable text classification}.
\newblock \emph{Proceedings of the AAAI Conference on Artificial Intelligence}, 35(15):13578--13586.

\bibitem[{OpenAI(2023)}]{openai2023gpt4}
OpenAI. 2023.
\newblock \href {http://arxiv.org/abs/2303.08774} {Gpt-4 technical report}.

\bibitem[{Pan et~al.(2010)Pan, Ni, Sun, Yang, and Chen}]{10.1145/1772690.1772767}
Sinno~Jialin Pan, Xiaochuan Ni, Jian-Tao Sun, Qiang Yang, and Zheng Chen. 2010.
\newblock \href {https://doi.org/10.1145/1772690.1772767} {Cross-domain sentiment classification via spectral feature alignment}.
\newblock In \emph{Proceedings of the 19th International Conference on World Wide Web}, WWW '10, page 751–760, New York, NY, USA. Association for Computing Machinery.

\bibitem[{Paszke et~al.(2019)Paszke, Gross, Massa, Lerer, Bradbury, Chanan, Killeen, Lin, Gimelshein, Antiga et~al.}]{paszke2019pytorch}
Adam Paszke, Sam Gross, Francisco Massa, Adam Lerer, James Bradbury, Gregory Chanan, Trevor Killeen, Zeming Lin, Natalia Gimelshein, Luca Antiga, et~al. 2019.
\newblock Pytorch: An imperative style, high-performance deep learning library.
\newblock \emph{Advances in neural information processing systems}, 32.

\bibitem[{Ribeiro et~al.(2020)Ribeiro, Wu, Guestrin, and Singh}]{ribeiro-etal-2020-beyond}
Marco~Tulio Ribeiro, Tongshuang Wu, Carlos Guestrin, and Sameer Singh. 2020.
\newblock \href {https://doi.org/10.18653/v1/2020.acl-main.442} {Beyond accuracy: Behavioral testing of {NLP} models with {C}heck{L}ist}.
\newblock In \emph{Proceedings of the 58th Annual Meeting of the Association for Computational Linguistics}, pages 4902--4912, Online. Association for Computational Linguistics.

\bibitem[{Saito and Saenko(2021)}]{saito2021ovanet}
Kuniaki Saito and Kate Saenko. 2021.
\newblock Ovanet: One-vs-all network for universal domain adaptation.
\newblock \emph{arXiv preprint arXiv:2104.03344}.

\bibitem[{Tack et~al.(2020)Tack, Mo, Jeong, and Shin}]{NEURIPS2020_8965f766}
Jihoon Tack, Sangwoo Mo, Jongheon Jeong, and Jinwoo Shin. 2020.
\newblock \href {https://proceedings.neurips.cc/paper/2020/file/8965f76632d7672e7d3cf29c87ecaa0c-Paper.pdf} {Csi: Novelty detection via contrastive learning on distributionally shifted instances}.
\newblock In \emph{Advances in Neural Information Processing Systems}, volume~33, pages 11839--11852. Curran Associates, Inc.

\bibitem[{Touvron et~al.(2023)Touvron, Lavril, Izacard, Martinet, Lachaux, Lacroix, Rozière, Goyal, Hambro, Azhar, Rodriguez, Joulin, Grave, and Lample}]{touvron2023llama}
Hugo Touvron, Thibaut Lavril, Gautier Izacard, Xavier Martinet, Marie-Anne Lachaux, Timothée Lacroix, Baptiste Rozière, Naman Goyal, Eric Hambro, Faisal Azhar, Aurelien Rodriguez, Armand Joulin, Edouard Grave, and Guillaume Lample. 2023.
\newblock \href {http://arxiv.org/abs/2302.13971} {Llama: Open and efficient foundation language models}.

\bibitem[{Wang et~al.(2023)Wang, Hu, Hou, Chen, Zheng, Wang, Yang, Huang, Ye, Geng, Jiao, Zhang, and Xie}]{wang2013robustness}
Jindong Wang, Xixu Hu, Wenxin Hou, Hao Chen, Runkai Zheng, Yidong Wang, Linyi Yang, Haojun Huang, Wei Ye, Xiubo Geng, Binxin Jiao, Yue Zhang, and Xing Xie. 2023.
\newblock On the robustness of chatgpt: An adversarial and out-of-distribution perspective.
\newblock \emph{arXiv preprint arXiv:2302.12095}.

\bibitem[{Wolf et~al.(2020)Wolf, Debut, Sanh, Chaumond, Delangue, Moi, Cistac, Rault, Louf, Funtowicz, Davison, Shleifer, von Platen, Ma, Jernite, Plu, Xu, Le~Scao, Gugger, Drame, Lhoest, and Rush}]{wolf-etal-2020-transformers}
Thomas Wolf, Lysandre Debut, Victor Sanh, Julien Chaumond, Clement Delangue, Anthony Moi, Pierric Cistac, Tim Rault, Remi Louf, Morgan Funtowicz, Joe Davison, Sam Shleifer, Patrick von Platen, Clara Ma, Yacine Jernite, Julien Plu, Canwen Xu, Teven Le~Scao, Sylvain Gugger, Mariama Drame, Quentin Lhoest, and Alexander Rush. 2020.
\newblock \href {https://doi.org/10.18653/v1/2020.emnlp-demos.6} {Transformers: State-of-the-art natural language processing}.
\newblock In \emph{Proceedings of the 2020 Conference on Empirical Methods in Natural Language Processing: System Demonstrations}, pages 38--45, Online. Association for Computational Linguistics.

\bibitem[{Wu and Shi(2022)}]{wu-shi-2022-adversarial}
Hui Wu and Xiaodong Shi. 2022.
\newblock \href {https://doi.org/10.18653/v1/2022.acl-long.174} {Adversarial soft prompt tuning for cross-domain sentiment analysis}.
\newblock In \emph{Proceedings of the 60th Annual Meeting of the Association for Computational Linguistics (Volume 1: Long Papers)}, pages 2438--2447, Dublin, Ireland. Association for Computational Linguistics.

\bibitem[{You et~al.(2019)You, Long, Cao, Wang, and Jordan}]{UDA_2019_CVPR}
Kaichao You, Mingsheng Long, Zhangjie Cao, Jianmin Wang, and Michael~I. Jordan. 2019.
\newblock Universal domain adaptation.
\newblock In \emph{The IEEE Conference on Computer Vision and Pattern Recognition (CVPR)}.

\bibitem[{Zeng et~al.(2021)Zeng, He, Yan, Liu, Wu, Xu, Jiang, and Xu}]{zeng-etal-2021-modeling}
Zhiyuan Zeng, Keqing He, Yuanmeng Yan, Zijun Liu, Yanan Wu, Hong Xu, Huixing Jiang, and Weiran Xu. 2021.
\newblock \href {https://doi.org/10.18653/v1/2021.acl-short.110} {Modeling discriminative representations for out-of-domain detection with supervised contrastive learning}.
\newblock In \emph{Proceedings of the 59th Annual Meeting of the Association for Computational Linguistics and the 11th International Joint Conference on Natural Language Processing (Volume 2: Short Papers)}, pages 870--878, Online. Association for Computational Linguistics.

\bibitem[{Zhan et~al.(2021)Zhan, Liang, Liu, Fan, Wu, and Lam}]{zhan-etal-2021-scope}
Li-Ming Zhan, Haowen Liang, Bo~Liu, Lu~Fan, Xiao-Ming Wu, and Albert~Y.S. Lam. 2021.
\newblock \href {https://doi.org/10.18653/v1/2021.acl-long.273} {Out-of-scope intent detection with self-supervision and discriminative training}.
\newblock In \emph{Proceedings of the 59th Annual Meeting of the Association for Computational Linguistics and the 11th International Joint Conference on Natural Language Processing (Volume 1: Long Papers)}, pages 3521--3532, Online. Association for Computational Linguistics.

\bibitem[{Zheng et~al.(2020)Zheng, Chen, and Huang}]{zheng2020out}
Yinhe Zheng, Guanyi Chen, and Minlie Huang. 2020.
\newblock Out-of-domain detection for natural language understanding in dialog systems.
\newblock \emph{IEEE/ACM Transactions on Audio, Speech, and Language Processing}, 28:1198--1209.

\end{thebibliography}
\bibliographystyle{acl_natbib}

\appendix

% \input{Assets/Tables/appendix_learning_rates_v2}

%%%%%%%%%%%%%%%%%%%%%%%%%%%%%%%%%%%%%%%%%%%%%%%%%%%%%%%%%%%%%%%%%%%%%%%%%%%%%%%%%%%%%%%%%%%%%%%%%%%%%%%%%%%%

\section{Full Experimental Results}
\label{sec:appendix_full_experimental_results}

\subsection{UniDA Results}
Table \ref{table:appendix_unida} is the full results of UniDA methods in our proposed testbed.
Baseline methods such as UAN \citep{UDA_2019_CVPR} and CMU \citep{Fu2020LearningTD} are included in the results.
We can observe that UniDA methods do not always retrain the same level of applicability in NLP.
Specifically, UAN and CMU utilize a fixed threshold defined in the vision domain.
While CMU remains fully compatible in the NLP domain, UAN struggles to apply effectively, as it fails to detect \textit{unknown} samples.

\subsection{CDA Results}
In this section, we demonstrate the experimental results of CDA methods with two additional scoring functions: cosine similarity and Mahalanobis distance.
The threshold value was selected based on the score from the scoring functions, using the same approach as the main experiment.
Also, we report the results of DANN \citep{Ganin2016DomainAdversarialTO} and source-only fine-tuning which was left out from the main experiment.
In some cases, source-only fine-tuning outperforms other adaptation methods, which is also observed in the vision domain \citep{UDA_2019_CVPR}.

\begin{enumerate}
    \item 
    \textbf{Cosine Similarity} \citep{NEURIPS2020_8965f766}
    calculates the cosine similarity score between the test input and the train input.
    The score is selected as the cosine similarity between the input and the nearest neighbor.
    The results are reported in Table \ref{table:appendix_cosine}.

    \item 
    \textbf{Mahalanobis Distance} \citep{NEURIPS2018_abdeb6f5}
    is the distance of the test sample to each class distribution.
    The representation is assumed to follow the multivariate normal distributions.
    The distance between the nearest class distribution is used as the score.
    The results are demonstrated in Table \ref{table:appendix_maha}
    
\end{enumerate}

Additionally, the full experimental results of MSP thresholding are presented in Table \ref{table:appendix_msp}.

%%%%%%%%%%%%%%%%%%%%%%%%%%%%%%%%%%%%%%%%%%%%%%%%%%%%%%%%%%%%%%%%%%%%%%%%%%%%%%%%%%%%%%%%%%%%%%%%%%%%%%%%%%%%

\begin{table}
    \centering
    \resizebox{0.99 \linewidth}{!}{
    
\begin{tabular}{c|cccc}
\toprule
CDA Methods & BERT & DANN & UDALM & AdSPT \\ \midrule
Huffpost (2013) & 5e-5 & 5e-5 & 1e-4 & 5e-5 \\
Huffpost (2014) & 5e-5 & 5e-5 & 1e-4 & 5e-5 \\
Huffpost (2015) & 5e-5 & 5e-5 & 1e-4 & 5e-5 \\
Huffpost (2016) & 5e-5 & 5e-5 & 1e-4 & 5e-5 \\
Huffpost (2017) & 5e-5 & 1e-4 & 1e-4 & 5e-5 \\
CLINC-150 & 1e-4 & 5e-4 & 5e-4 & 1e-5 \\
MASSIVE & 5e-5 & 5e-5 & 5e-5 & 5e-5 \\
Amazon & 1e-5 & 5e-6 & 5e-5 & 1e-5 \\ \midrule
UniDA Methods & UAN & CMU & OVANet & UniOT \\ \midrule
Huffpost (2013) & 5e-5 & 5e-5 & 5e-5 & 1e-4 \\
Huffpost (2014) & 5e-5 & 5e-5 & 5e-5 & 1e-4 \\
Huffpost (2015) & 1e-4 & 1e-4 & 5e-5 & 5e-5 \\
Huffpost (2016) & 1e-4 & 5e-5 & 5e-5 & 1e-4 \\
Huffpost (2017) & 1e-4 & 5e-5 & 5e-5 & 1e-4 \\
CLINC-150 & 5e-6 & 1e-5 & 5e-6 & 5e-5 \\
MASSIVE & 5e-5 & 5e-5 & 5e-5 & 1e-4 \\
Amazon & 5e-5 & 5e-5 & 5e-5 & 5e-5 \\ 
\bottomrule
\end{tabular}

    }
    \caption{
        \label{table:appendix_learning_rates}
        Learning rates for each methods.
        The learning rate was selected from 5e-4, 1e-4, 5e-5, 1e-5, and 1e-6 with the best evaluation performance from the source domain.
    }
\end{table}

\begin{table*}[t]
    \centering    
    \resizebox{0.99 \textwidth}{!}{

\begin{tabular}{c|ccc|ccc}
\toprule
Dataset & \multicolumn{3}{c|}{Huffpost (2012 $\rightarrow$ 2013) (3 / 4 / 4)} & \multicolumn{3}{c}{Huffpost (2012 $\rightarrow$ 2014) (3 / 4 / 4)} \\ \midrule
Method & $acc_{C}$ & $acc_{\bar{C}_t}$ & H-score & $acc_{C}$ & $acc_{\bar{C}_t}$ & H-score \\ \midrule
UAN & 73.58 {\footnotesize $\pm 5.76$} & 0.00 {\footnotesize $\pm 0.00$} & 0.00 {\footnotesize $\pm 0.00$} & 36.38 {\footnotesize $\pm 3.04$} & 0.00 {\footnotesize $\pm 0.00$} & 0.00 {\footnotesize $\pm 0.00$} \\
CMU & 58.18 {\footnotesize $\pm 1.88$} & 26.89 {\footnotesize $\pm 4.29$} & \underline{36.76} {\footnotesize $\pm 4.22$} & 33.13 {\footnotesize $\pm 3.35$} & 39.98 {\footnotesize $\pm 3.71$} & 36.02 {\footnotesize $\pm 1.46$} \\
OVANet & 65.11 {\footnotesize $\pm 0.60$} & 24.91 {\footnotesize $\pm 6.75$} & 35.64 {\footnotesize $\pm 6.70$} & 45.85 {\footnotesize $\pm 2.62$} & 33.96 {\footnotesize $\pm 6.78$} & \underline{38.56} {\footnotesize $\pm 3.95$} \\
UniOT & 53.76 {\footnotesize $\pm 1.10$} & 65.86 {\footnotesize $\pm 3.57$} & \textbf{59.14} {\footnotesize $\pm 1.14$} & 33.49 {\footnotesize $\pm 4.79$} & 71.44 {\footnotesize $\pm 6.19$} & \textbf{45.23} {\footnotesize $\pm 3.60$} \\ \midrule
Dataset & \multicolumn{3}{c|}{Huffpost (2012 $\rightarrow$ 2015) (3 / 4 / 4)} & \multicolumn{3}{c}{Huffpost (2012 $\rightarrow$ 2016) (3 / 4 / 4)} \\ \midrule
Method & $acc_{C}$ & $acc_{\bar{C}_t}$ & H-score & $acc_{C}$ & $acc_{\bar{C}_t}$ & H-score \\ \midrule
UAN & 34.17 {\footnotesize $\pm 6.00$} & 0.00 {\footnotesize $\pm 0.00$} & 0.00 {\footnotesize $\pm 0.00$} & 42.53 {\footnotesize $\pm 16.52$} & 0.00 {\footnotesize $\pm 0.00$} & 0.00 {\footnotesize $\pm 0.00$} \\
CMU & 39.57 {\footnotesize $\pm 3.54$} & 17.98 {\footnotesize $\pm 6.30$} & 24.02 {\footnotesize $\pm 5.57$} & 41.80 {\footnotesize $\pm 2.08$} & 30.09 {\footnotesize $\pm 3.51$} & 34.84 {\footnotesize $\pm 1.97$} \\
OVANet & 43.91 {\footnotesize $\pm 2.09$} & 35.70 {\footnotesize $\pm 4.41$} & \underline{39.21} {\footnotesize $\pm 2.43$} & 44.30 {\footnotesize $\pm 2.89$} & 33.49 {\footnotesize $\pm 4.79$} & \underline{38.03} {\footnotesize $\pm 3.79$} \\
UniOT & 27.16 {\footnotesize $\pm 3.40$} & 75.50 {\footnotesize $\pm 2.68$} & \textbf{39.82} {\footnotesize $\pm 3.39$} & 33.09 {\footnotesize $\pm 3.62$} & 69.27 {\footnotesize $\pm 3.83$} & \textbf{44.60} {\footnotesize $\pm 2.85$} \\ \midrule
Dataset & \multicolumn{3}{c|}{Huffpost (2012 $\rightarrow$ 2017) (3 / 4 / 4)} & \multicolumn{3}{c}{CLINC-150 (4 / 3 / 3)} \\ \midrule
Method & $acc_{C}$ & $acc_{\bar{C}_t}$ & H-score & $acc_{C}$ & $acc_{\bar{C}_t}$ & H-score \\ \midrule
UAN & 22.74 {\footnotesize $\pm 10.63$} & 0.00 {\footnotesize $\pm 0.00$} & 0.00 {\footnotesize $\pm 0.00$} & 48.99 {\footnotesize $\pm 14.49$} & 0.00 {\footnotesize $\pm 0.00$} & 0.00 {\footnotesize $\pm 0.00$} \\
CMU & 34.87 {\footnotesize $\pm 3.14$} & 55.79 {\footnotesize $\pm 2.71$} & \underline{42.86} {\footnotesize $\pm 2.72$} & 60.70 {\footnotesize $\pm 1.02$} & 36.89 {\footnotesize $\pm 1.52$} & \underline{45.87} {\footnotesize $\pm 1.09$} \\
OVANet & 48.37 {\footnotesize $\pm 4.12$} & 35.03 {\footnotesize $\pm 3.60$} & 40.46 {\footnotesize $\pm 2.32$} & 83.49 {\footnotesize $\pm 0.62$} & 31.24 {\footnotesize $\pm 1.70$} & 45.45 {\footnotesize $\pm 1.88$} \\
UniOT & 31.57 {\footnotesize $\pm 1.93$} & 75.85 {\footnotesize $\pm 9.63$} & \textbf{44.36} {\footnotesize $\pm 1.27$} & 64.14 {\footnotesize $\pm 9.14$} & 77.36 {\footnotesize $\pm 3.94$} & \textbf{69.88} {\footnotesize $\pm 6.11$} \\ \midrule
Dataset & \multicolumn{3}{c|}{MASSIVE (8 / 5 / 5)} & \multicolumn{3}{c}{Amazon (11 / 10 / 10)} \\ \midrule
Method & $acc_{C}$ & $acc_{\bar{C}_t}$ & H-score & $acc_{C}$ & $acc_{\bar{C}_t}$ & H-score \\ \midrule
UAN & 13.00 {\footnotesize $\pm 7.18$} & 0.00 {\footnotesize $\pm 0.00$} & 0.00 {\footnotesize $\pm 0.00$} & 35.40 {\footnotesize $\pm 3.44$} & 0.00 {\footnotesize $\pm 0.00$} & 0.00 {\footnotesize $\pm 0.00$} \\
CMU & 37.38 {\footnotesize $\pm 1.59$} & 10.33 {\footnotesize $\pm 3.06$} & 16.03 {\footnotesize $\pm 3.84$} & 44.30 {\footnotesize $\pm 0.90$} & 39.95 {\footnotesize $\pm 9.67$} & 41.48 {\footnotesize $\pm 5.10$} \\
OVANet & 59.04 {\footnotesize $\pm 0.69$} & 39.25 {\footnotesize $\pm 2.14$} & \underline{47.12} {\footnotesize $\pm 1.47$} & 44.50 {\footnotesize $\pm 0.84$} & 42.49 {\footnotesize $\pm 3.95$} & \underline{43.38} {\footnotesize $\pm 1.83$} \\
UniOT & 44.01 {\footnotesize $\pm 3.52$} & 79.19 {\footnotesize $\pm 4.14$} & \textbf{56.52} {\footnotesize $\pm 3.49$} & 38.60 {\footnotesize $\pm 1.08$} & 70.98 {\footnotesize $\pm 7.76$} & \textbf{49.90} {\footnotesize $\pm 2.08$} \\ 
\bottomrule
\end{tabular}

    }
    \caption{
        Experimental results of UniDA methods in the proposed testbed.
        For each dataset, the \textbf{best method} with the highest H-score is in bold and the \underline{second-best method} is underlined.
    }
    \label{table:appendix_unida}
\end{table*}

\begin{table*}[t]
    \centering    
    \resizebox{0.99 \textwidth}{!}{

  \begin{tabular}{c|cccccc}
\toprule
Dataset & \multicolumn{3}{c|}{Huffpost (2012 $\rightarrow$ 2013) (3 / 4 / 4)} & \multicolumn{3}{c}{Huffpost (2012 $\rightarrow$ 2014) (3 / 4 / 4)} \\ \midrule
Method & $acc_{C}$ & $acc_{\bar{C}_t}$ & \multicolumn{1}{c|}{H-score} & $acc_{C}$ & $acc_{\bar{C}_t}$ & H-score \\ \midrule
BERT & 64.06 {\footnotesize $\pm 2.13$} & 28.02 {\footnotesize $\pm 3.03$} & \multicolumn{1}{c|}{\textbf{38.86} {\footnotesize $\pm 2.63$}} & 40.63 {\footnotesize $\pm 3.55$} & 37.41 {\footnotesize $\pm 4.81$} & \textbf{38.65} {\footnotesize $\pm 1.34$} \\
DANN & 58.92 {\footnotesize $\pm 1.62$} & 1.69 {\footnotesize $\pm 0.57$} & \multicolumn{1}{c|}{3.28 {\footnotesize $\pm 1.07$}} & 33.08 {\footnotesize $\pm 2.94$} & 2.01 {\footnotesize $\pm 0.82$} & 3.76 {\footnotesize $\pm 1.48$} \\
UDALM & 60.12 {\footnotesize $\pm 2.56$} & 23.94 {\footnotesize $\pm 3.80$} & \multicolumn{1}{c|}{\underline{34.12} {\footnotesize $\pm 4.00$}} & 37.97 {\footnotesize $\pm 1.69$} & 29.27 {\footnotesize $\pm 6.15$} & \underline{32.84} {\footnotesize $\pm 4.22$} \\
AdSPT & 62.71 {\footnotesize $\pm 0.88$} & 10.89 {\footnotesize $\pm 3.04$} & \multicolumn{1}{c|}{18.41 {\footnotesize $\pm 4.38$}} & 39.57 {\footnotesize $\pm 3.81$} & 23.88 {\footnotesize $\pm 3.44$} & 29.74 {\footnotesize $\pm 3.59$} \\ \midrule
Dataset & \multicolumn{3}{c}{Huffpost (2012 $\rightarrow$ 2015) (3 / 4 / 4)} & \multicolumn{3}{c}{Huffpost (2012 $\rightarrow$ 2016) (3 / 4 / 4)} \\ \midrule
Method & $acc_{C}$ & $acc_{\bar{C}_t}$ & \multicolumn{1}{c|}{H-score} & $acc_{C}$ & $acc_{\bar{C}_t}$ & H-score \\ \midrule
BERT & 40.75 {\footnotesize $\pm 2.12$} & 30.90 {\footnotesize $\pm 1.55$} & \multicolumn{1}{c|}{\textbf{35.13} {\footnotesize $\pm 1.53$}} & 47.07 {\footnotesize $\pm 2.46$} & 26.15 {\footnotesize $\pm 3.73$} & \textbf{33.42} {\footnotesize $\pm 2.43$} \\
DANN & 31.71 {\footnotesize $\pm 2.49$} & 3.39 {\footnotesize $\pm 0.92$} & \multicolumn{1}{c|}{6.11 {\footnotesize $\pm 1.57$}} & 36.68 {\footnotesize $\pm 3.86$} & 2.16 {\footnotesize $\pm 0.80$} & 4.05 {\footnotesize $\pm 1.44$} \\
UDALM & 37.06 {\footnotesize $\pm 1.74$} & 24.79 {\footnotesize $\pm 4.83$} & \multicolumn{1}{c|}{\underline{29.59} {\footnotesize $\pm 3.89$}} & 44.10 {\footnotesize $\pm 1.52$} & 23.26 {\footnotesize $\pm 2.48$} & \underline{30.38} {\footnotesize $\pm 2.02$} \\
AdSPT & 39.92 {\footnotesize $\pm 4.75$} & 15.01 {\footnotesize $\pm 3.85$} & \multicolumn{1}{c|}{21.74 {\footnotesize $\pm 4.84$}} & 41.44 {\footnotesize $\pm 3.63$} & 12.66 {\footnotesize $\pm 3.31$} & 19.14 {\footnotesize $\pm 3.97$} \\ \midrule
Dataset & \multicolumn{3}{c}{Huffpost (2012 $\rightarrow$ 2017) (3 / 4 / 4)} & \multicolumn{3}{c}{CLINC-150 (4 / 3 / 3)} \\ \midrule
Method & $acc_{C}$ & $acc_{\bar{C}_t}$ & \multicolumn{1}{c|}{H-score} & $acc_{C}$ & $acc_{\bar{C}_t}$ & H-score \\ \midrule
BERT & 48.84 {\footnotesize $\pm 3.16$} & 29.52 {\footnotesize $\pm 4.67$} & \multicolumn{1}{c|}{\textbf{36.62} {\footnotesize $\pm 3.91$}} & 74.53 {\footnotesize $\pm 3.02$} & 71.37 {\footnotesize $\pm 6.54$} & \underline{72.69} {\footnotesize $\pm 2.48$} \\
DANN & 31.30 {\footnotesize $\pm 2.34$} & 9.32 {\footnotesize $\pm 1.63$} & \multicolumn{1}{c|}{14.31 {\footnotesize $\pm 2.02$}} & 60.21 {\footnotesize $\pm 9.33$} & 49.65 {\footnotesize $\pm 14.72$} & 52.74 {\footnotesize $\pm 10.15$} \\
UDALM & 41.08 {\footnotesize $\pm 3.92$} & 29.10 {\footnotesize $\pm 7.01$} & \multicolumn{1}{c|}{33.48 {\footnotesize $\pm 3.83$}} & 75.99 {\footnotesize $\pm 3.63$} & 78.11 {\footnotesize $\pm 5.41$} & \textbf{76.90} {\footnotesize $\pm 2.35$} \\
AdSPT & 43.02 {\footnotesize $\pm 0.74$} & 31.07 {\footnotesize $\pm 4.77$} & \multicolumn{1}{c|}{\underline{35.93} {\footnotesize $\pm 3.37$}} & 42.94 {\footnotesize $\pm 5.59$} & 6.96 {\footnotesize $\pm 3.44$} & 11.57 {\footnotesize $\pm 4.59$} \\ \midrule
Dataset & \multicolumn{3}{c|}{MASSIVE (8 / 5 / 5)} & \multicolumn{3}{c}{Amazon (11 / 10 / 10)} \\ \midrule
Method & $acc_{C}$ & $acc_{\bar{C}_t}$ & \multicolumn{1}{c|}{H-score} & $acc_{C}$ & $acc_{\bar{C}_t}$ & H-score \\ \midrule
BERT & 51.21 {\footnotesize $\pm 3.62$} & 62.96 {\footnotesize $\pm 2.06$} & \multicolumn{1}{c|}{\underline{56.44} {\footnotesize $\pm 2.69$}} & 44.88 {\footnotesize $\pm 2.10$} & 14.49 {\footnotesize $\pm 1.25$} & \underline{21.90} {\footnotesize $\pm 1.68$} \\
DANN & 41.55 {\footnotesize $\pm 9.00$} & 13.74 {\footnotesize $\pm 3.52$} & \multicolumn{1}{c|}{20.62 {\footnotesize $\pm 5.04$}} & 44.58 {\footnotesize $\pm 0.97$} & 19.61 {\footnotesize $\pm 1.08$} & \textbf{27.21} {\footnotesize $\pm 0.93$} \\
UDALM & 65.22 {\footnotesize $\pm 2.67$} & 64.25 {\footnotesize $\pm 2.17$} & \multicolumn{1}{c|}{\textbf{64.67} {\footnotesize $\pm 0.76$}} & 46.65 {\footnotesize $\pm 0.47$} & 13.58 {\footnotesize $\pm 2.23$} & 20.96 {\footnotesize $\pm 2.79$} \\
AdSPT & 36.78 {\footnotesize $\pm 2.65$} & 59.87 {\footnotesize $\pm 13.47$} & \multicolumn{1}{c|}{45.32 {\footnotesize $\pm 6.20$}} & 30.26 {\footnotesize $\pm 1.45$} & 6.45 {\footnotesize $\pm 4.55$} & 10.15 {\footnotesize $\pm 5.65$} \\ 
\bottomrule
\end{tabular}

    }
    \caption{
        Experimental results of CDA methods with \textit{cosine similarity} as the scoring function.
        For each dataset, the \textbf{best method} with the highest H-score is in bold and the \underline{second-best method} is underlined.
    }
    \label{table:appendix_cosine}
\end{table*}

\begin{table*}[t]
    \centering    
    \resizebox{0.99 \textwidth}{!}{

  \begin{tabular}{c|ccc|ccc}
\toprule
Dataset & \multicolumn{3}{c|}{Huffpost (2012 $\rightarrow$ 2013) (3 / 4 / 4)} & \multicolumn{3}{c}{Huffpost (2012 $\rightarrow$ 2014) (3 / 4 / 4)} \\ \midrule
Method & $acc_{C}$ & $acc_{\bar{C}_t}$ & H-score & $acc_{C}$ & $acc_{\bar{C}_t}$ & H-score \\ \midrule
BERT & 10.40 {\footnotesize $\pm 12.11$} & 35.69 {\footnotesize $\pm 5.72$} & \textbf{13.78} {\footnotesize $\pm 12.38$} & 11.28 {\footnotesize $\pm 8.86$} & 42.44 {\footnotesize $\pm 2.47$} & \underline{16.41} {\footnotesize $\pm 9.26$} \\
DANN & 18.55 {\footnotesize $\pm 9.65$} & 1.50 {\footnotesize $\pm 0.30$} & 2.69 {\footnotesize $\pm 0.62$} & 16.78 {\footnotesize $\pm 7.67$} & 1.48 {\footnotesize $\pm 0.67$} & 2.70 {\footnotesize $\pm 1.22$} \\
UDALM & 7.39 {\footnotesize $\pm 10.12$} & 27.73 {\footnotesize $\pm 4.25$} & 9.26 {\footnotesize $\pm 10.18$} & 4.92 {\footnotesize $\pm 4.14$} & 34.04 {\footnotesize $\pm 5.66$} & 8.15 {\footnotesize $\pm 6.52$} \\
AdSPT & 18.07 {\footnotesize $\pm 10.66$} & 7.70 {\footnotesize $\pm 0.36$} & \underline{10.21} {\footnotesize $\pm 2.49$} & 17.29 {\footnotesize $\pm 8.06$} & 18.80 {\footnotesize $\pm 3.47$} & \textbf{16.87} {\footnotesize $\pm 5.39$} \\ \midrule
Dataset & \multicolumn{3}{c|}{Huffpost (2012 $\rightarrow$ 2015) (3 / 4 / 4)} & \multicolumn{3}{c}{Huffpost (2012 $\rightarrow$ 2016) (3 / 4 / 4)} \\ \midrule
Method & $acc_{C}$ & $acc_{\bar{C}_t}$ & H-score & $acc_{C}$ & $acc_{\bar{C}_t}$ & H-score \\ \midrule
BERT & 11.28 {\footnotesize $\pm 11.10$} & 37.71 {\footnotesize $\pm 4.18$} & \underline{15.38} {\footnotesize $\pm 11.27$} & 10.34 {\footnotesize $\pm 9.85$} & 31.43 {\footnotesize $\pm 5.76$} & \textbf{14.52} {\footnotesize $\pm 10.93$} \\
DANN & 20.09 {\footnotesize $\pm 5.97$} & 2.93 {\footnotesize $\pm 1.23$} & 4.88 {\footnotesize $\pm 1.59$} & 20.79 {\footnotesize $\pm 10.71$} & 1.92 {\footnotesize $\pm 0.32$} & 3.44 {\footnotesize $\pm 0.41$} \\
UDALM & 4.19 {\footnotesize $\pm 2.98$} & 31.19 {\footnotesize $\pm 6.15$} & 7.06 {\footnotesize $\pm 4.84$} & 6.23 {\footnotesize $\pm 5.09$} & 31.52 {\footnotesize $\pm 5.19$} & 9.49 {\footnotesize $\pm 6.74$} \\
AdSPT & 19.20 {\footnotesize $\pm 7.31$} & 14.55 {\footnotesize $\pm 3.02$} & \textbf{16.30} {\footnotesize $\pm 4.47$} & 17.78 {\footnotesize $\pm 9.51$} & 9.73 {\footnotesize $\pm 4.76$} & \underline{12.10} {\footnotesize $\pm 6.48$} \\ \midrule
Dataset & \multicolumn{3}{c|}{Huffpost (2012 $\rightarrow$ 2017) (3 / 4 / 4)} & \multicolumn{3}{c}{CLINC-150 (4 / 3 / 3)} \\ \midrule
Method & $acc_{C}$ & $acc_{\bar{C}_t}$ & H-score & $acc_{C}$ & $acc_{\bar{C}_t}$ & H-score \\ \midrule
BERT & 10.31 {\footnotesize $\pm 13.20 $} & 34.04 {\footnotesize $\pm 5.04 $} & \underline{13.10} {\footnotesize $\pm 13.18$} & 16.67 {\footnotesize $\pm 21.48$} & 72.56 {\footnotesize $\pm 5.44$} & \textbf{21.87} {\footnotesize $\pm 25.69$} \\
DANN & 16.79 {\footnotesize $\pm 0.94$} & 7.48 {\footnotesize $\pm 1.62$} & 10.29 {\footnotesize $\pm 1.63$} & 1.87 {\footnotesize $\pm 1.36$} & 41.89 {\footnotesize $\pm 20.06$} & 3.25 {\footnotesize $\pm 1.89$} \\
UDALM & 7.65 {\footnotesize $\pm 5.29$} & 30.08 {\footnotesize $\pm 8.33$} & 10.93 {\footnotesize $\pm 6.62$} & 5.18 {\footnotesize $\pm 8.80$} & 80.15 {\footnotesize $\pm 3.95$} & \underline{8.73} {\footnotesize $\pm 14.40$} \\
AdSPT & 19.77 {\footnotesize $\pm 8.68$} & 30.93 {\footnotesize $\pm 3.52$} & \textbf{23.30} {\footnotesize $\pm 7.35$} & 6.59 {\footnotesize $\pm 7.97$} & 29.85 {\footnotesize $\pm 29.28$} & 7.41{\footnotesize $\pm 5.75$} \\ \midrule
Dataset & \multicolumn{3}{c|}{MASSIVE (8 / 5 / 5)} & \multicolumn{3}{c}{Amazon (11 / 10 / 10)} \\ \midrule
Method & $acc_{C}$ & $acc_{\bar{C}_t}$ & H-score & $acc_{C}$ & $acc_{\bar{C}_t}$ & H-score \\ \midrule
BERT & 5.62 {\footnotesize $\pm 9.41$} & 68.53 {\footnotesize $\pm 4.54$} & \underline{9.07} {\footnotesize $\pm 14.57$} & 4.66 {\footnotesize $\pm 2.59$} & 15.97 {\footnotesize $\pm 1.59$} & \underline{6.85} {\footnotesize $\pm 3.45$} \\
DANN & 4.59 {\footnotesize $\pm 3.59$} & 9.73 {\footnotesize $\pm 3.60$} & 5.26 {\footnotesize $\pm 2.70$} & 8.62 {\footnotesize $\pm 7.31$} & 17.74 {\footnotesize $\pm 3.69$} & \textbf{9.75} {\footnotesize $\pm 3.92$} \\
UDALM & 2.96 {\footnotesize $\pm 4.90$} & 62.80 {\footnotesize $\pm 5.02$} & 5.17 {\footnotesize $\pm 8.36$} & 2.68 {\footnotesize $\pm 2.80$} & 19.01 {\footnotesize $\pm 5.49$} & 4.48 {\footnotesize $\pm 4.22$} \\
AdSPT & 5.70 {\footnotesize $\pm 6.41$} & 54.61 {\footnotesize $\pm 12.44$} & \textbf{9.60} {\footnotesize $\pm 10.75$} & 4.63 {\footnotesize $\pm 4.66$} & 9.38 {\footnotesize $\pm 6.68$} & 5.61 {\footnotesize $\pm 5.65$} \\ 
\bottomrule
\end{tabular}

    }
    \caption{
        Experimental results of CDA methods with \textit{Mahalanobis distance} as the scoring function.
        For each dataset, the \textbf{best method} with the highest H-score is in bold and the \underline{second-best method} is underlined.
    }
    \label{table:appendix_maha}
\end{table*}

\begin{table*}[t]
    \centering    
    \resizebox{0.99 \textwidth}{!}{

\begin{tabular}{c|ccc|ccc}
\toprule
Dataset & \multicolumn{3}{c|}{Huffpost (2012 $\rightarrow$ 2013) (3 / 4 / 4)} & \multicolumn{3}{c}{Huffpost (2012 $\rightarrow$ 2014) (3 / 4 / 4)} \\ \midrule
Method & $acc_{C}$ & $acc_{\bar{C}_t}$ & H-score & $acc_{C}$ & $acc_{\bar{C}_t}$ & H-score \\ \midrule
BERT & 51.47 {\footnotesize $\pm 3.72$} & 72.07 {\footnotesize $\pm 10.30$} & \underline{59.64} {\footnotesize $\pm 3.28$} & 26.71 {\footnotesize $\pm 1.99$} & 78.98 {\footnotesize $\pm 8.23$} & \underline{39.75} {\footnotesize $\pm 1.51$} \\
DANN & 49.35 {\footnotesize $\pm 6.58$} & 77.79 {\footnotesize $\pm 25.05$} & 57.83 {\footnotesize $\pm 6.49$} & 14.94 {\footnotesize $\pm 3.26$} & 95.24 {\footnotesize $\pm 0.70$} & 25.71 {\footnotesize $\pm 4.48$} \\
UDALM & 52.74 {\footnotesize $\pm 2.92$} & 58.25 {\footnotesize $\pm 6.03$} & 55.15 {\footnotesize $\pm 2.37$} & 29.90 {\footnotesize $\pm 1.89$} & 50.00 {\footnotesize $\pm 17.49$} & 36.14 {\footnotesize $\pm 5.07$} \\
AdSPT & 55.05 {\footnotesize $\pm 2.11$} & 80.66 {\footnotesize $\pm 2.99$} & \textbf{65.38} {\footnotesize $\pm 1.00$} & 29.64 {\footnotesize $\pm 4.51$} & 80.03 {\footnotesize $\pm 6.09$} & \textbf{42.93} {\footnotesize $\pm 3.71$} \\ \midrule
Dataset & \multicolumn{3}{c|}{Huffpost (2012 $\rightarrow$ 2015) (3 / 4 / 4)} & \multicolumn{3}{c}{Huffpost (2012 $\rightarrow$ 2016) (3 / 4 / 4)} \\ \midrule
Method & $acc_{C}$ & $acc_{\bar{C}_t}$ & H-score & $acc_{C}$ & $acc_{\bar{C}_t}$ & H-score \\ \midrule
BERT & 27.35 {\footnotesize $\pm 2.64$} & 77.13 {\footnotesize $\pm 2.20$} & \underline{40.30} {\footnotesize $\pm 2.60$} & 33.12 {\footnotesize $\pm 1.85$} & 63.65 {\footnotesize $\pm 3.63$} & \textbf{43.53} {\footnotesize $\pm 1.96$} \\
DANN & 16.48 {\footnotesize $\pm 7.90$} & 89.93 {\footnotesize $\pm 11.51$} & 26.81 {\footnotesize $\pm 9.33$} & 11.23 {\footnotesize $\pm 4.13$} & 95.41 {\footnotesize $\pm 1.23$} & 19.89 {\footnotesize $\pm 6.52$} \\
UDALM & 25.15 {\footnotesize $\pm 2.60$} & 61.79 {\footnotesize $\pm 8.88$} & 35.56 {\footnotesize $\pm 3.15$} & 29.14 {\footnotesize $\pm 1.59$} & 60.87 {\footnotesize $\pm 6.87$} & 39.28 {\footnotesize $\pm 1.44$} \\
AdSPT & 29.08 {\footnotesize $\pm 5.39$} & 73.04 {\footnotesize $\pm 12.71$} & \textbf{40.78} {\footnotesize $\pm 3.42$} & 28.84 {\footnotesize $\pm 3.57$} & 82.52 {\footnotesize $\pm 4.22$} & \underline{42.55} {\footnotesize $\pm 3.27$} \\ \midrule
Dataset & \multicolumn{3}{c|}{Huffpost (2012 $\rightarrow$ 2017) (3 / 4 / 4)} & \multicolumn{3}{c}{CLINC-150 (4 / 3 / 3)} \\ \midrule
Method & $acc_{C}$ & $acc_{\bar{C}_t}$ & H-score & $acc_{C}$ & $acc_{\bar{C}_t}$ & H-score \\ \midrule
BERT & 33.07 {\footnotesize $\pm 3.90$} & 65.82 {\footnotesize $\pm 7.47$} & \textbf{43.64} {\footnotesize $\pm 1.71$} & 64.69 {\footnotesize $\pm 3.68$} & 72.08 {\footnotesize $\pm 9.82$} & \underline{67.82} {\footnotesize $\pm 4.15$} \\
DANN & 22.84 {\footnotesize $\pm 2.60$} & 82.46 {\footnotesize $\pm 4.39$} & 35.66 {\footnotesize $\pm 2.85$} & 48.02 {\footnotesize $\pm 4.93$} & 67.48 {\footnotesize $\pm 19.86$} & 54.90 {\footnotesize $\pm 9.17$} \\
UDALM & 32.85 {\footnotesize $\pm 1.05$} & 43.08 {\footnotesize $\pm 6.76$} & 37.08 {\footnotesize $\pm 2.82$} & 74.92 {\footnotesize $\pm 6.11$} & 69.91 {\footnotesize $\pm 13.98$} & \textbf{71.28} {\footnotesize $\pm 4.62$} \\
AdSPT & 35.74 {\footnotesize $\pm 3.33$} & 53.04 {\footnotesize $\pm 16.02$} & \underline{41.36} {\footnotesize $\pm 5.04$} & 3.96 {\footnotesize $\pm 2.17$} & 98.69 {\footnotesize $\pm 0.36$} & 7.55 {\footnotesize $\pm 3.95$} \\ \midrule
Dataset & \multicolumn{3}{c|}{MASSIVE (8 / 5 / 5)} & \multicolumn{3}{c}{Amazon (11 / 10 / 10)} \\ \midrule
Method & $acc_{C}$ & $acc_{\bar{C}_t}$ & H-score & $acc_{C}$ & $acc_{\bar{C}_t}$ & H-score \\ \midrule
BERT & 30.04 {\footnotesize $\pm 2.74$} & 68.95 {\footnotesize $\pm 5.53$} & \underline{41.79} {\footnotesize $\pm 3.20$} & 44.98 {\footnotesize $\pm 2.32$} & 12.66 {\footnotesize $\pm 1.47$} & 19.70 {\footnotesize $\pm 1.69$} \\
DANN & 20.37 {\footnotesize $\pm 11.32$} & 88.69 {\footnotesize $\pm 5.27$} & 31.77 {\footnotesize $\pm 13.32$} & 45.85 {\footnotesize $\pm 0.49$} & 12.52 {\footnotesize $\pm 0.71$} & 19.66 {\footnotesize $\pm 0.87$} \\
UDALM & 38.33 {\footnotesize $\pm 4.88$} & 41.29 {\footnotesize $\pm 15.14$} & 39.28 {\footnotesize $\pm 9.69$} & 47.69 {\footnotesize $\pm 1.47$} & 15.10 {\footnotesize $\pm 2.30$} & \underline{22.85} {\footnotesize $\pm 2.65$} \\
AdSPT & 36.78 {\footnotesize $\pm 2.65$} & 59.87 {\footnotesize $\pm 13.47$} & \textbf{45.32} {\footnotesize $\pm 6.20$} & 30.26 {\footnotesize $\pm 3.52$} & 30.93 {\footnotesize $\pm 26.14$} & \textbf{27.12} {\footnotesize $\pm 11.03$} \\ 
\bottomrule
\end{tabular}

    }
    \caption{
        Experimental results of CDA methods with \textit{Maximum Softmax Probability} as the scoring function.
        For each dataset, the \textbf{best method} with the highest H-score is in bold and the \underline{second-best method} is underlined.
    }
    \label{table:appendix_msp}
\end{table*}

%%%%%%%%%%%%%%%%%%%%%%%%%%%%%%%%%%%%%%%%%%%%%%%%%%%%%%%%%%%%%%%%%%%%%%%%%%%%%%%%%%%%%%%%%%%%%%%%%%%%%%%%%%%%
\section{Implementation Details}
\label{sec:implementation_details}
For the experiments, we adopt a 12-layer pre-trained language model \textit{bert-base-uncased} \citep{devlin-etal-2019-bert} as the backbone of all the methods.
We utilized the \texttt{[CLS]} representation as the input feature.
AdamW optimizer \citep{Loshchilov2019DecoupledWD} was used for all the experiments with a batch size of 32.
We selected the best learning rate among 5e-4, 1e-4, 5e-5, 1e-5, and 5e-6.
The learning rate for each method is reported in Table \ref{table:appendix_learning_rates}.
The model was trained for 10 epochs with an early stopping on the accuracy of the source domain's evaluation set.
All the experiments were implemented with Pytorch \citep{paszke2019pytorch} and Huggingface Transformers library \citep{wolf-etal-2020-transformers}.
The experiments take an hour on a single Tesla V100 GPU.

% \input{Assets/Figures/appendix_different_thresholds}
% \section{H-score Performance on Various Threshold Values}
% \label{sec:appendinx_different_thresholds}
% Figure \ref{figure:appendix_different_thresholds} illustrates the performance variations of all the CDA methods in our proposed testbed.
% The performance changes considerably depending on the threshold values, and in certain circumstances, they outperform UniDA methods.
% Note that finding the optimal threshold value without supervision is very limited, so the results must be considered for analysis purpose only.

%%%%%%%%%%%%%%%%%%%%%%%%%%%%%%%%%%%%%%%%%%%%%%%%%%%%%%%%%%%%%%%%%%%%%%%%%%%%%%%%%%%%%%%%%%%%%%%%%%%%%%%%%%%%

\begin{table*}[t]
    \centering    
    % \resizebox{0.9 \textwidth}{!}{

\begin{tabular}{c|ccc|ccc}
\toprule
Dataset & \multicolumn{3}{c|}{CLINC-150 (4 / 3 / 3)} & \multicolumn{3}{c}{MASSIVE (8 / 5 / 5)} \\ \midrule
Runs & 1 & 2 & 3 & 1 & 2 & 3 \\ \midrule
BERT & 44.59 {\footnotesize $\pm 6.01$} & 48.08 {\footnotesize $\pm 2.90$} & 52.25 {\footnotesize $\pm 3.54$} & 51.68 {\footnotesize $\pm 1.62$} & 46.68 {\footnotesize $\pm 1.90$} & \underline{51.56} {\footnotesize $\pm 0.81$} \\
DANN & 47.91 {\footnotesize $\pm 9.96$} & 53.23 {\footnotesize $\pm 7.92$} & \underline{60.42} {\footnotesize $\pm 7.00$} & 47.16 {\footnotesize $\pm 3.79$} & 31.39 {\footnotesize $\pm 3.93$} & 45.21 {\footnotesize $\pm 8.93$} \\
UDALM & \underline{49.17} {\footnotesize $\pm 8.75$} & \underline{55.04} {\footnotesize $\pm 5.31$} & 52.42 {\footnotesize $\pm 8.20$} & 55.52 {\footnotesize $\pm 3.00$} & 42.34 {\footnotesize $\pm 5.88$} & 51.15 {\footnotesize $\pm 3.62$} \\
AdSPT & 31.81  {\footnotesize $\pm 8.63$} & 48.62 {\footnotesize $\pm 1.85$} & 46.20 {\footnotesize $\pm 5.22$} & 45.41 {\footnotesize $\pm 9.04$} & 38.61 {\footnotesize $\pm 3.90$} & 46.95 {\footnotesize $\pm 5.22$} \\ \midrule
UAN & 0.00 {\footnotesize $\pm 0.00$} & 0.00 {\footnotesize $\pm 0.00$} & 0.00 {\footnotesize $\pm 0.00$} & 0.00 {\footnotesize $\pm 0.00$} & 0.00 {\footnotesize $\pm 0.00$} & 0.00 {\footnotesize $\pm 0.00$} \\
CMU & 40.32 {\footnotesize $\pm 2.32$} & 38.79 {\footnotesize $\pm 2.19$} & 40.92 {\footnotesize $\pm 1.77$} & 21.46 {\footnotesize $\pm 1.62$} & 18.84 {\footnotesize $\pm 5.26$} & 28.33 {\footnotesize $\pm 4.29$} \\
OVANet & 41.77 {\footnotesize $\pm 1.90$} & 35.89 {\footnotesize $\pm 3.86$} & 29.15 {\footnotesize $\pm 2.17$} & \textbf{63.91} {\footnotesize $\pm 3.41$} & \underline{54.31} {\footnotesize $\pm 0.63$} & 39.85 {\footnotesize $\pm 3.74$} \\
UniOT & \textbf{50.94} {\footnotesize $\pm 7.20$} & \textbf{61.66} {\footnotesize $\pm 1.70$} & \underline{56.51} {\footnotesize $\pm 3.53$} & \underline{62.07} {\footnotesize $\pm 2.89$} & \textbf{55.11} {\footnotesize $\pm 2.31$} & \textbf{63.59} {\footnotesize $\pm 3.15$} \\ \bottomrule
\end{tabular}

    % }
    \caption{
        H-score results of UniDA methods in CLINC-150 and MASSIVE with different class splits.
        The \textbf{best method} with the highest H-score is in bold and the \underline{second-best method} is underlined.
    }
    \label{table:appendix_different_split}
\end{table*}
\section{Ablation on Different Class Splits}
\label{sec:appendix_different_splits}
For the main experiment, we utilized class names as the criterion to implement the category gap. 
However, this may only show the specific scenario of the category gap.
To provide a more comprehensive analysis, we also report the results when the class set is randomly split.
We utilized CLINC-150 and MASSIVE dataset for the ablation study, and MSP thresholding was applied for CDA methods.
We conducted three experiments, each with a different class split, and for every split, we reported the average results of three different runs.

Table \ref{table:appendix_different_split} is the results of the experiments.
Due to the changes in the class set to be predicted, the task difficulty varies, resulting in differences in absolute performances.
However, when comparing the relative performance between different methods, 
we can observe that they exhibit consistent trends regardless of the class split.

\section{Receiver Operating Characteristic (ROC) Curve}
\begin{figure*}[t]
    \centering
    \begin{subfigure}{0.23\textwidth}
        \includegraphics[width=\textwidth]{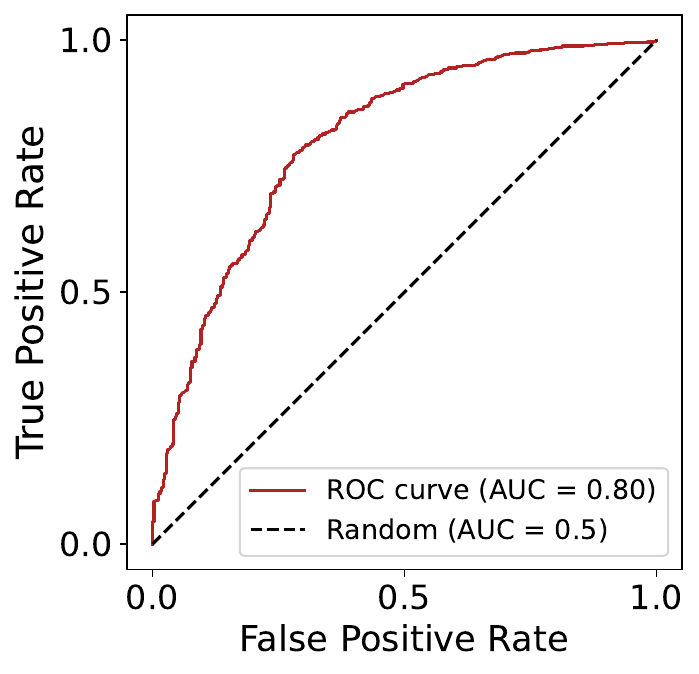}
        \caption{Huffpost (2013)}
    \end{subfigure}
    \hfill
    \begin{subfigure}{0.245\textwidth}
        \includegraphics[width=\textwidth]{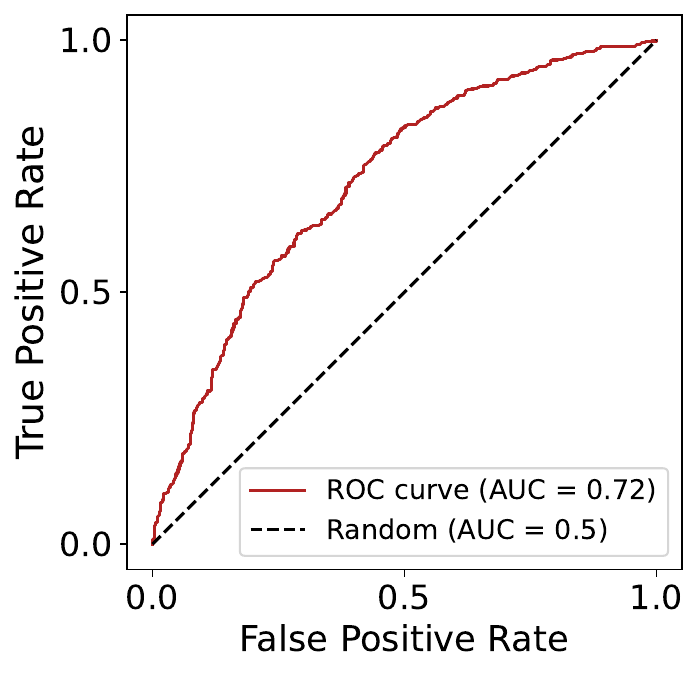}
        \caption{Huffpost (2014)}
    \end{subfigure}
    \hfill
    \begin{subfigure}{0.245\textwidth}
        \includegraphics[width=\textwidth]{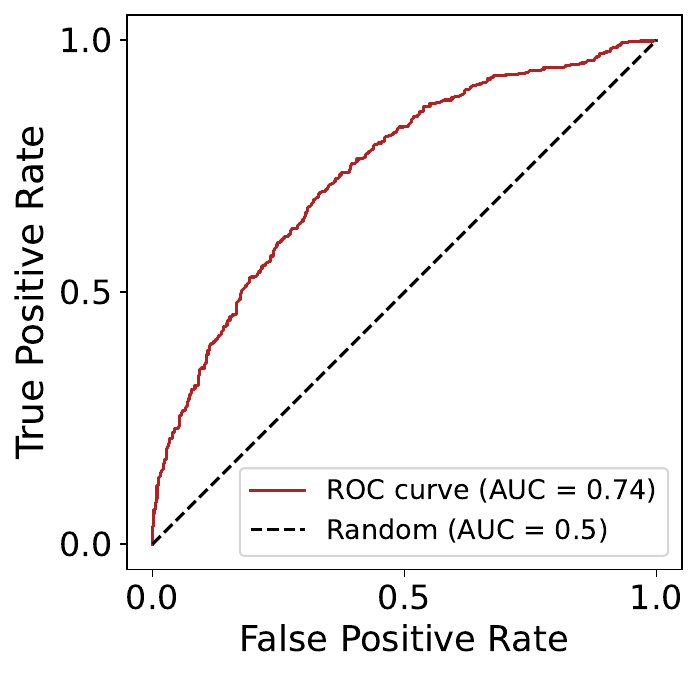}
        \caption{Huffpost (2015)}
    \end{subfigure}
    \hfill
    \begin{subfigure}{0.245\textwidth}
        \includegraphics[width=\textwidth]{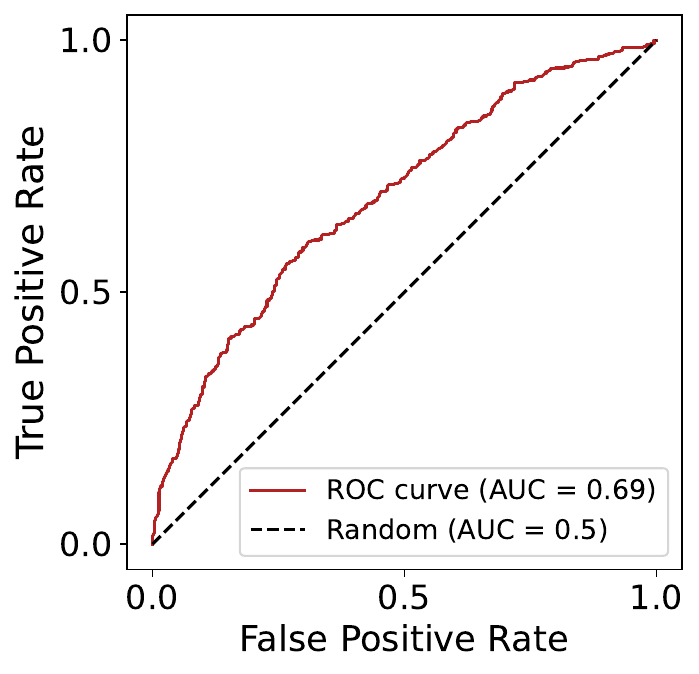}
        \caption{Huffpost (2016)}
    \end{subfigure}
    \hfill
    \begin{subfigure}{0.245\textwidth}
        \includegraphics[width=\textwidth]{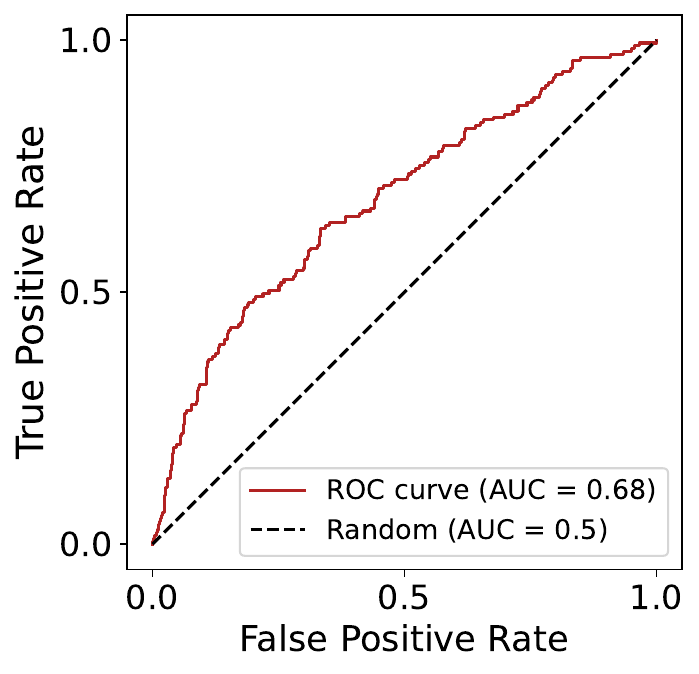}
        \caption{Huffpost (2017)}
    \end{subfigure}
    \hfill
    \begin{subfigure}{0.245\textwidth}
        \includegraphics[width=\textwidth]{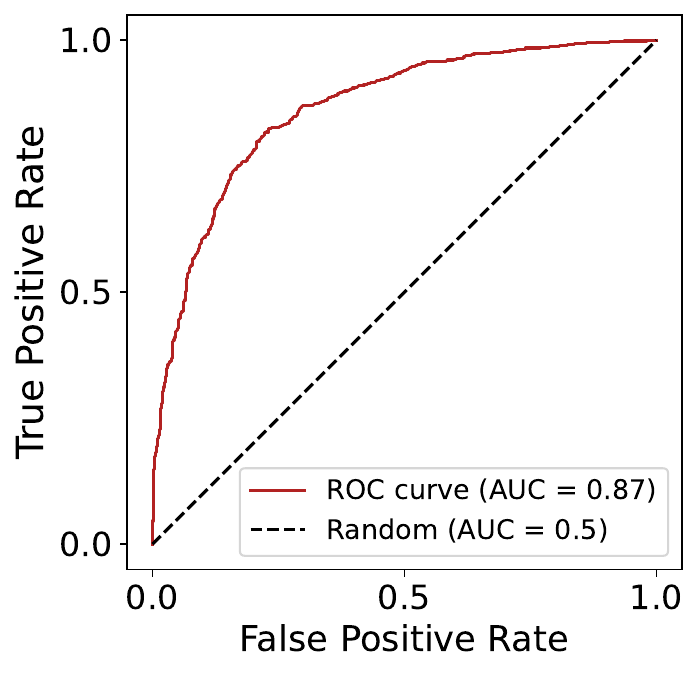}
        \caption{CLINC-150}
    \end{subfigure}
    \hfill
    \begin{subfigure}{0.245\textwidth}
        \includegraphics[width=\textwidth]{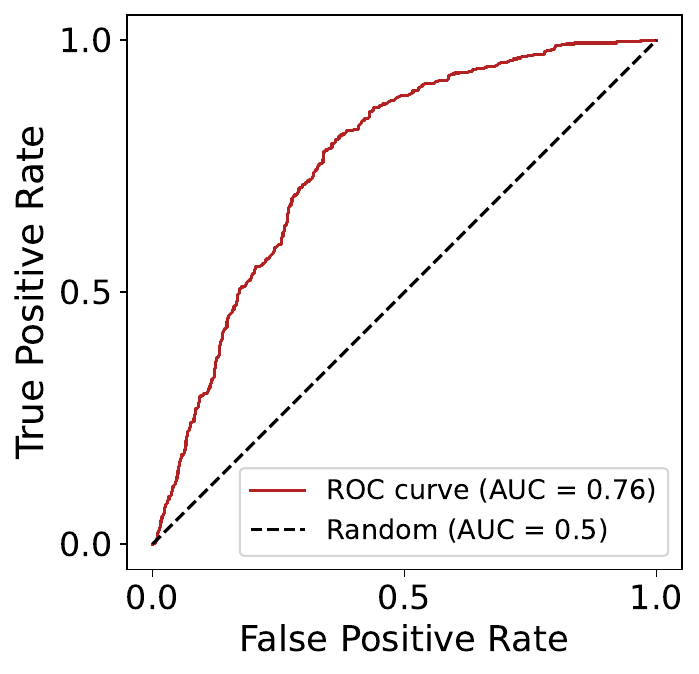}
        \caption{MASSIVE}
    \end{subfigure}
    \hfill
    \begin{subfigure}{0.245\textwidth}
        \includegraphics[width=\textwidth]{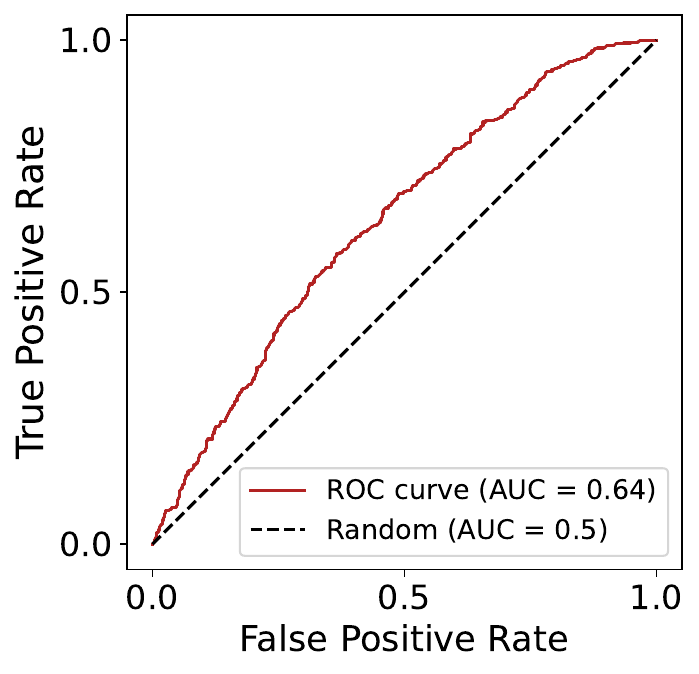}
        \caption{Amazon}
    \end{subfigure}

    \caption{
        ROC curve of discerning \textit{unknown} samples from the \textit{transferable} samples.
        The closer the ROC curve is to the upper-left corner, it becomes easier to distinguish between the them.
    }
    \label{figure:appendix_roc}
\end{figure*}

To measure Distinction Difficulty Score (DDS), we calculated the AUROC and subtracted from 1.
Figure \ref{figure:appendix_roc} is the ROC curve of discerning \textit{unknown} inputs from the \textit{transferable} inputs 
for our proposed datasets.
The closer the ROC curve is to the upper-left corner, 
it indicates that it is easier to distinguish between \textit{unknown} and \textit{transferable} inputs.

% \input{Assets/Figures/appendix_different_thresholds}

% \section{Various Threshold Values}
% We present H-score visualization of the full datasets of our testbed.
% The results are visualized in Figure \ref{figure:appendix_different_thresholds}.
% In easy or moderate adaptation scenario, 
% such as CLINC-150 and Huffpost(2012, 2013, 2014, 2015),
% CDA exhibit the potential to outperform UniDA methods. 
% However, in difficult adaptations, such as Amazon, MASSIVE, and Huffpost (2017),
% UniDA methods significantly outperforms CDA methods.
% Even with an optimal threshold, CDA methods cannot surpass UniDA methods.

\end{document}